\documentclass[runningheads]{llncs}

\usepackage{eccv}

\usepackage{eccvabbrv}

\usepackage{graphicx}
\usepackage{booktabs}

\usepackage[accsupp]{axessibility}  

\usepackage{hyperref}

\usepackage{orcidlink}

\usepackage[most]{tcolorbox}
\usepackage{graphicx}
\usepackage{amsmath,amssymb} 
\usepackage{color}
\usepackage{xcolor}
\usepackage{tabularx}
\usepackage{booktabs}
\usepackage{caption}
\usepackage{lipsum} 
\usepackage{multirow}
\usepackage{xurl}
\usepackage{hyperref}  
\usepackage{arydshln}
\usepackage{pifont}
\newcommand{\cmark}{\ding{51}}  
\newcommand{\xmark}{\ding{55}}  

\tcbset{
  mybox/.style={
    enhanced,
    colframe=gray!30,
    colback=white,
    coltitle=black,
    title=Editing Pair Construction,
    fonttitle=\bfseries\small,
    top=10pt,
    bottom=10pt,
    left=10pt,
    right=10pt,
    boxsep=5pt,
    arc=8pt, 
    outer arc=8pt,
    boxrule=0.5pt,
    before skip=10pt,
    after skip=10pt,
    overlay={\node[anchor=north west, inner sep=2pt] at (frame.north west) {\textcolor{black}{\tiny\textbullet}};},
    attach boxed title to top center={yshift=-2mm,yshifttext=-2mm},
    boxed title style={colback=gray!30, colframe=gray!30, fonttitle=\bfseries\small, boxsep=2pt, boxrule=0.5pt, sharp corners},
  },
}

\begin{document}

\title{SIQA: Toward Reliable Scientific Image Quality Assessment}

\titlerunning{SIQA: Toward Reliable Scientific Image Quality Assessment}

\author{
Wenzhe Li\inst{1*} \and 
Liang Chen\inst{2,3*} \and
Junying Wang\inst{2} \and
Yijing Guo \inst{2} \and
Ye Shen \inst{2} \and
Farong Wen \inst{2} \and
Chunyi Li \inst{2} \and
Zicheng Zhang \inst{2\dag} \and
Guangtao Zhai \inst{2,3\dag}
}
\authorrunning{Wenzhe Li and Liang Chen~Author et al.}

\institute{
TongJi University \and
Shanghai Artificial Intelligence Laboratory \and
Shanghai Jiao Tong University 
}

\maketitle

\begin{abstract}

Scientific images fundamentally differ from natural and AI-generated images in that they encode structured domain knowledge rather than merely depict visual scenes. Assessing their quality therefore requires evaluating not only perceptual fidelity but also scientific correctness and logical completeness. However, existing image quality assessment (IQA) paradigms primarily focus on perceptual distortions or image–text alignment, implicitly assuming that depicted content is factually valid. This assumption breaks down in scientific contexts, where visually plausible figures may still contain conceptual errors or incomplete reasoning.
To address this gap, we introduce \textbf{Scientific Image Quality Assessment (SIQA)}, a framework that models scientific image quality along two complementary dimensions: \textbf{Knowledge} (\textit{Scientific Validity} and \textit{Scientific Completeness}) and \textbf{Perception} (\textit{Cognitive Clarity} and \textit{Disciplinary Conformity}). To operationalize this formulation, we design two evaluation protocols: \textbf{SIQA-U} (Understanding), which measures semantic comprehension of scientific content through multiple-choice tasks, and \textbf{SIQA-S} (Scoring), which evaluates alignment with expert quality judgments. We further construct the \textbf{SIQA Challenge}, consisting of an expert-annotated benchmark and a large-scale training set.
Experiments across representative multimodal large language models (MLLMs) reveal a consistent discrepancy between scoring alignment and scientific understanding. While models can achieve strong agreement with expert ratings under SIQA-S, their performance on SIQA-U remains substantially lower. Fine-tuning improves both metrics, yet gains in scoring consistently outpace improvements in understanding. These results suggest that rating consistency alone may not reliably reflect scientific comprehension, underscoring the necessity of multidimensional evaluation for scientific image quality assessment.

\end{abstract}

\section{Introduction}

\begin{figure}[t]
    \centering
    \includegraphics[width=0.8\linewidth]{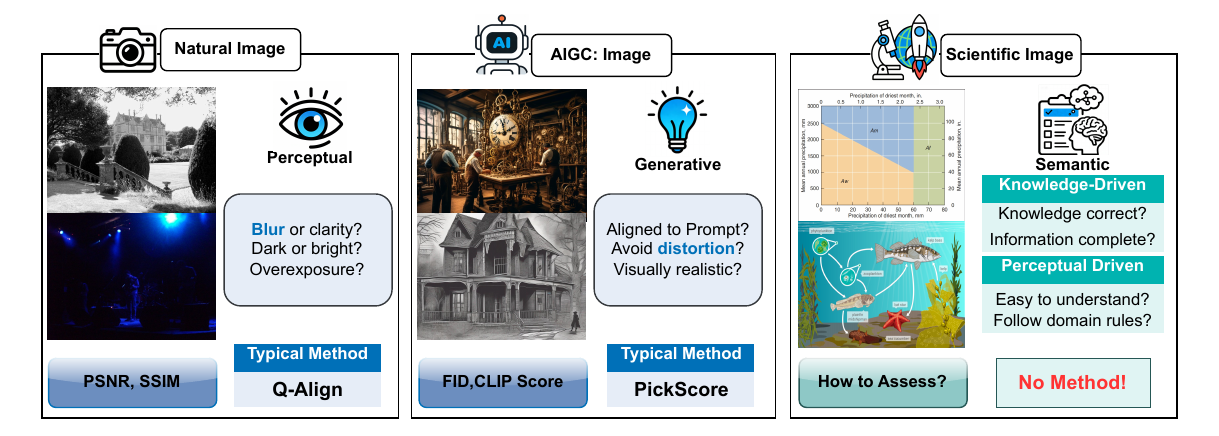}
    \caption{
    Comparison of quality assessment methods across image domains. Traditional IQA focuses on perceptual fidelity (e.g., PSNR, SSIM), while AIGC evaluation emphasizes image-text alignment (e.g., FID, CLIP Score). In contrast, scientific images require joint evaluation of perceptual clarity and semantic correctness, but no dedicated framework currently exists.}
    \label{fig:diff}
    \vspace{-1.5em}
\end{figure}

Scientific images, such as molecular structures, reaction schematics, and geometric diagrams, are structured visual representations of domain knowledge rather than mere depictions of physical scenes. Unlike natural photographs, whose quality is typically assessed in terms of perceptual fidelity or aesthetic coherence, scientific figures must satisfy both visual interpretability and scientific correctness. A figure may appear visually polished yet remain scientifically flawed due to factual inconsistencies, logical omissions, or violations of disciplinary conventions; thus, perceptual realism alone is insufficient as a proxy for quality.
Existing image quality assessment (IQA) frameworks implicitly assume that image content is factually correct and therefore concentrate on perceptual distortions (e.g., blur, noise) or image–text alignment. While such assumptions are reasonable for natural images, which rarely require verification against structured scientific knowledge, they become insufficient for scientific figures. As illustrated in Fig.~\ref{fig:diff}, current paradigms evaluate perceptual fidelity or instruction alignment, yet none explicitly examine whether the conveyed content adheres to scientific facts or disciplinary conventions. 
We therefore argue that scientific image quality cannot be adequately characterized along a single perceptual axis. Instead, it requires joint consideration of perceptual fidelity and knowledge correctness. Existing IQA frameworks capture only the former, leaving the epistemic dimension of scientific imagery largely unexplored.

To formalize this two-dimensional perspective, we ground scientific image quality in established principles of scientific practice and visualization theory. In scientific reasoning, content evaluation is governed by reproducibility and logical inference~\cite{national2019reproducibility}, which imply two essential criteria: \textit{scientific validity}, referring to factual consistency with established knowledge, and \textit{scientific completeness}, requiring the inclusion of all necessary elements for sound inference. 
Complementarily, the design of scientific figures follows domain-specific conventions and cognitive principles. Authoritative guidelines mandate adherence to \textit{disciplinary conformity} (e.g., IUPAC chemical notation~\cite{Brecher+2008+277+410}), while effective visual communication emphasizes \textit{cognitive clarity} through coherent layout, consistent labeling, and reduced cognitive load during interpretation~\cite{MPS}. 
These principles imply that scientific image quality cannot be reduced to perceptual attributes alone, but must explicitly encode epistemic correctness and communicative structure.

Building on these principles, we propose \textbf{Scientific Image Quality Assessment (SIQA)}, a framework organized along two complementary dimensions: \textbf{Knowledge} (\textit{Scientific Validity and Scientific Completeness}) and \textbf{Perception} (\textit{Cognitive Clarity and Disciplinary Conformity}).
Beyond defining quality dimensions, evaluation methodology itself presents a challenge. Conventional IQA protocols typically rely on holistic quality scoring, which may conflate semantic understanding with rating alignment. To disentangle these factors, we introduce two complementary evaluation protocols. \textbf{SIQA-U} (Understanding) measures semantic comprehension of scientific content through multiple-choice reasoning tasks grounded in the proposed dimensions. \textbf{SIQA-S} (Scoring) evaluates alignment with expert quality judgments. This separation enables us to examine whether strong scoring performance truly implies deep scientific comprehension.

\begin{table}[t]
\centering
\caption{Comparison of datasets across image domains, understanding tasks, and quality scoring. Perc. is perception while Know. is knowledge.}

\label{tab:dataset_comparison}
\resizebox{\columnwidth}{!}{%
\begin{tabular}{lcccccc}
\toprule
\textbf{Dataset} & \textbf{Domain} & \textbf{Understand} & \textbf{Questions Type} & \textbf{Score} & \textbf{Ratings Type} \\
\midrule
LIVE Challenge~\cite{ghadiyaram2015massive}    & NSI     & \xmark & —                 & \cmark & MOS                  \\
SPAQ ~\cite{fang2020perceptual}            & NSI      & \xmark & —                 & \cmark & MOS                  \\
CGIQA-6k~\cite{zhang2023subjective}       & CGI       & \xmark & —                 & \cmark & MOS                  \\
AGIQA-3k~\cite{li2023agiqa}         & AIGC            & \xmark & —                 & \cmark & MOS                  \\
\hdashline
GenExam ~\cite{wang2025genexam}          & AIGC            & \cmark & Knowledge QA      & \xmark & —                    \\
ChemVLM ~\cite{li2025chemvlm}          & Chemistry       & \cmark & Scientific Reasoning & \xmark & —                    \\
GeoTrust~\cite{fu2025trustgeogen}          & Geometry        & \cmark & Scientific Reasoning & \xmark & —                    \\
ScienceQA~\cite{lu2022learn}         & Scientific      & \cmark & MCQ               & \xmark & —                    \\
\hline

\textbf{SIQA Challenge(Ours)} & \textbf{Mixed}   & \textbf{\cmark} & \textbf{Quality-Centric MCQ} & \textbf{\cmark} & \textbf{MOS (Perc. \& Know.)} \\
\bottomrule
\end{tabular}
}
\vspace{-1.5em}
\end{table}

To enable empirical investigation of this framework, we introduce the \textbf{SIQA Challenge}, consisting of an expert-annotated benchmark and a large-scale training set aligned with the proposed dimensions. Using the SIQA-U and SIQA-S protocols, we evaluate representative multimodal large language models (MLLMs) and further fine-tune an open-source model to obtain \textbf{SIQA-Judger}. 
Across models, we observe a clear discrepancy between scoring alignment and scientific understanding: agreement with expert ratings under SIQA-S is generally strong, whereas performance on the reasoning-oriented SIQA-U tasks is comparatively lower. Fine-tuning improves results under both protocols, yet scoring performance improves more rapidly, whereas gains in scientific understanding remain comparatively limited.
This systematic discrepancy suggests that rating alignment alone may not reliably reflect scientific comprehension, highlighting the importance of multidimensional evaluation in scientific image quality assessment.

The main contributions of this work are:
\begin{itemize}
    \item We introduce \textbf{SIQA}, the first framework that explicitly models scientific image quality with knowledge (Scientific Validity and Scientific Completeness) and perception (Cognitive Clarity and Disciplinary Conformity).
    \item We construct the \textbf{SIQA Challenge}, an expert-annotated benchmark and large-scale training set that explicitly disentangles semantic understanding (SIQA-U) from rating alignment (SIQA-S).
    \item Through comprehensive evaluation of mainstream MLLMs and a fine-tuned model (\textbf{SIQA-Judger}), we empirically demonstrate a consistent decoupling between scoring agreement and scientific understanding.
\end{itemize}

\begin{figure}[t]
    \centering
    \includegraphics[width=1\linewidth]{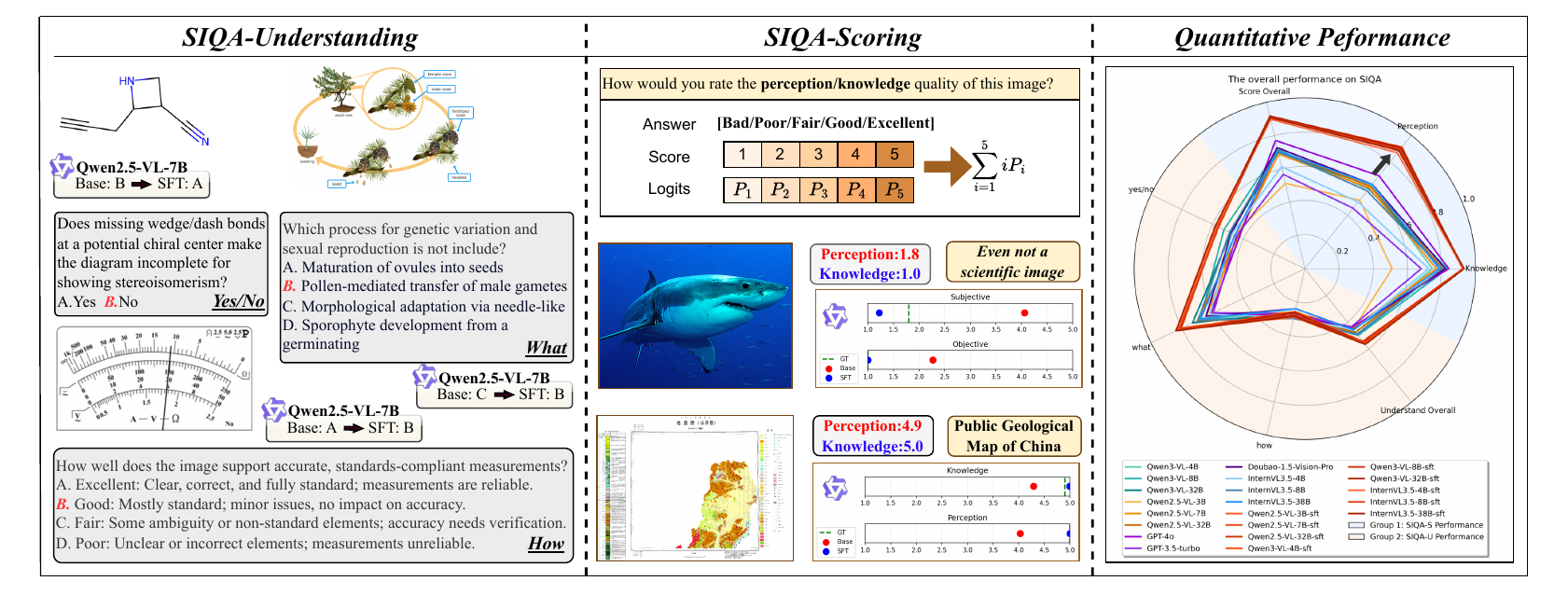}
    \caption{Quantitative and qualitative evaluation of \textsc{SIQA} on multiple models. Left and Middle: Case studies illustrating \textsc{SIQA-U} and \textsc{SIQA-S} protocols across question types. Right: Radar map showing performance gains from fine-tuning.}
    \label{fig:spotlight}
    \vspace{-1.5em}
\end{figure}

\section{Related Work}

\subsection{MLLMs for Scientific Understanding}
MLLMs have evolved from contrastive alignment frameworks such as CLIP to instruction-tuned architectures like LLaVA~\cite{liu2023llava,liu2023improvedllava}, enabling strong performance on open-domain vision–language tasks including visual question answering and image captioning. 
Building on these advances, recent efforts have extended MLLMs to scientific domains. 
Models such as \textit{ChemDFM}~\cite{zhao2024chemdfm} and \textit{Mol-Instructions}~\cite{cao2025instructmol} specialize in chemistry and biomedicine through curated image–text instruction data, while systems like \textit{Kosmos-2.5}~\cite{lv2023kosmos} and \textit{PaperQA}~\cite{lala2023paperqa} aim to parse scientific figures or extract structured knowledge from research documents.

Despite their domain adaptation, these models are primarily designed for content recognition, reasoning, and information extraction. Their evaluation relies on task-specific metrics such as accuracy or BLEU score, which measure the correctness of generated answers rather than the scientific validity of the input image itself. Consequently, they implicitly assume that scientific images are factually correct and internally consistent. However, scientific image validity assessment constitutes a distinct problem. The ability to answer questions about an image does not guarantee the ability to determine whether the image itself is scientifically valid or complete. To date, the capacity of MLLMs to judge scientific image quality has not been systematically studied.

\subsection{Image Quality Assessment}

Image quality assessment (IQA) has primarily targeted natural scene images (NSIs) and computer-generated images (CGIs), focusing on technical distortions such as blur, noise, and compression artifacts that impair human perception. 
Early full-reference approaches~\cite{ML-IQA1,ML-IQA2} relied on metrics like PSNR and SSIM, which measure fidelity through pixel-wise error or structural similarity.
With the development of deep learning, no-reference IQA (NR-IQA) models such as HyperIQA~\cite{CNN-IQA1} and WaDIQaM~\cite{CNN-IQA2} emerged, predicting perceptual quality end-to-end using hierarchical visual features learned from Mean Opinion Score (MOS) datasets. This evolution reflects a gradual shift from low-level signal fidelity to modeling human perceptual preference. More recently, Transformer-based vision language foundation models such as CLIP~\cite{CLIP} and its variants CLIP-IQA~\cite{CLIPIQA} evaluate image quality by measuring semantic consistency between visual content and textual prompts.
Concurrently, MLLMs~\cite{zhang2025large,Q-Instruct} have demonstrated advanced cross-modal reasoning capabilities. Compared to NSIs and CGI, quality evaluation of AIGC~\cite{ghosh2023geneval} further emphasizes alignment with user instructions, incorporating high-level semantic reasoning into the notion of quality.

However, prevailing IQA frameworks are fundamentally centered on fidelity to a reference signal, human judgment, or user intent, rather than the factual correctness of the depicted content.  
This limitation is especially pronounced in scientific images, where quality is defined not only by perceptual fidelity but also by the scientific correctness, logical consistency, and empirical verifiability of the content.
Consequently, a significant gap exists between current IQA frameworks and the requirements of scientific image evaluation, revealing an urgent need for approaches that explicitly assess the scientific validity of visual content.

\begin{figure}[t]
    \centering
    \includegraphics[width=1\linewidth]{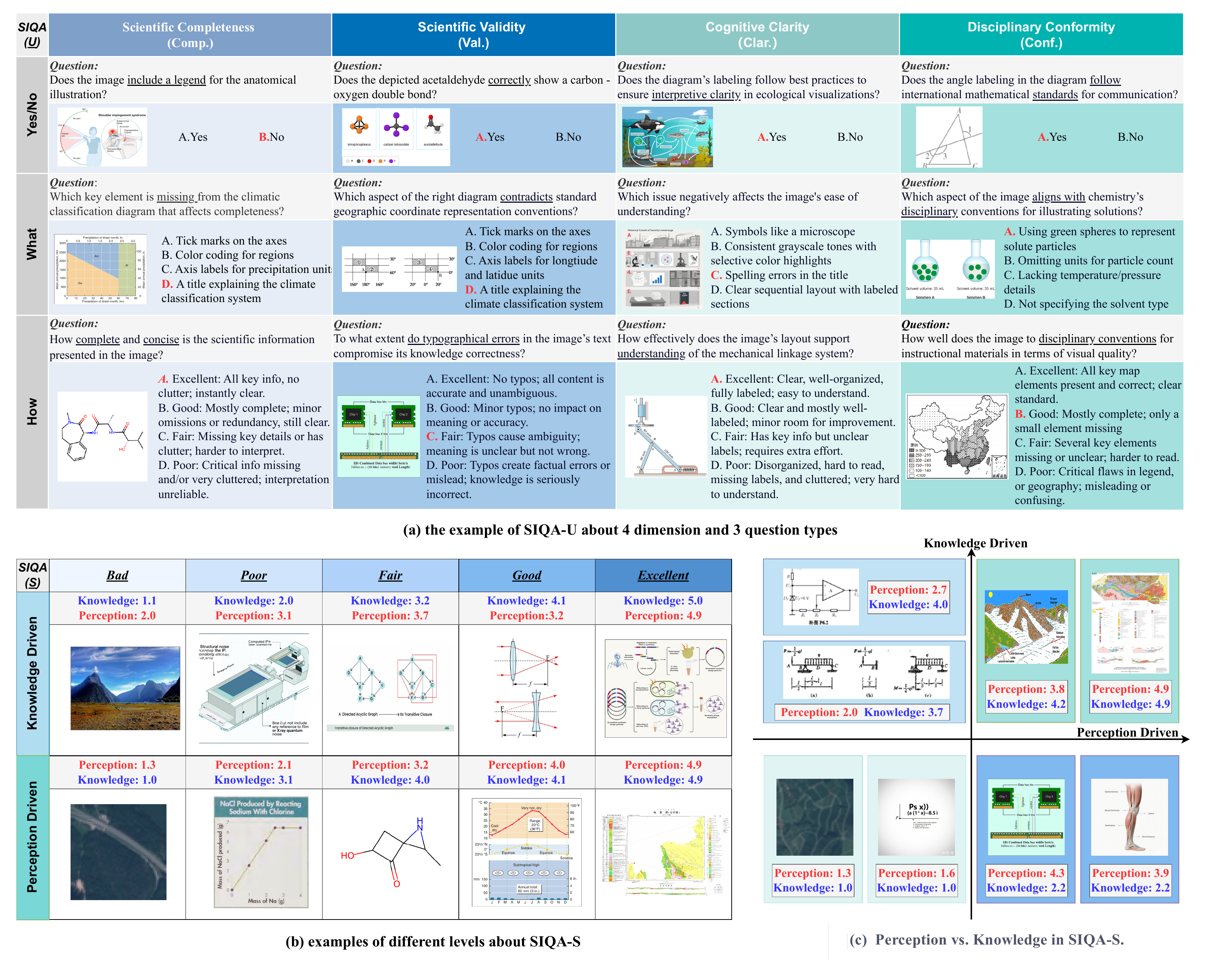}
    \caption{
    Examples from the SIQA Challenge, illustrating its design and annotation scheme. (a) Sample questions from SIQA-U across four dimensions. (b) Example of different levels in SIQA-S. (c) Comparison of Perception and Knowledge scores across image-quality levels, illustrating their partial correlation and structural distinction.
    }
    \label{fig:dataset_visual}
    \vspace{-1.5em}
\end{figure}

\section{Constructing the SIQA Challenge}

This section outlines the process of constructing the SIQA Challenge. The guiding principles behind its design are first described (Sec.~\ref{Sec:principle}), followed by the data collection and question generation procedures (Sec.~\ref{sec:data_collection}). The expert annotation protocol is then detailed (Sec.~~\ref{Sec:annotation}), an analysis of the dataset’s statistical properties is provided (Sec.~~\ref{Sec:anaysis}), and the construction of supervised fine-tuning data is presented (Sec.~\ref{Sec:model SFT}).

\subsection{Design Principles}
\label{Sec:principle}
\paragraph{\textbf{Covering dimensions of SIQA}}
Scientific images impose stringent semantic requirements on the content they convey, rendering conventional IQA frameworks based on low-level visual attributes inadequate. 
SIQA models scientific image quality through two complementary dimensions: 
(1) perception, encompassing Cognitive Clarity and Disciplinary Conformity, 
and (2) knowledge, capturing Scientific Validity and Scientific Completeness. 
Unlike conventional IQA frameworks focused on low-level attributes, SIQA emphasizes both cognitive interpretability and scientific grounding. 
The examples are provided in Fig.~\ref{fig:dataset_visual}.

\paragraph{\textbf{Ensuring Diversity}}
To account for the diverse application scenarios and broad disciplinary coverage of scientific imagery, images are aggregated from multiple scientific multimodal datasets, thereby maximizing the richness and representativeness of the SIQA Challenge dataset. 
Furthermore, to enhance the semantic and functional diversity of the collected scientific images, coarse-grained categorization is performed based on the scientific purpose, and balanced sampling is applied.
During question construction, MLLMs are employed at different stages, guided by category-specific prompts, to generate a wide spectrum of scientifically grounded questions. 
In evaluation, both open-source and closed-source MLLMs are tested. Collectively, these strategies ensure the diversity, robustness, and generalizability of the SIQA Challenge. Additional details about data collection are provided in Section~\ref{sec:data_collection}.

\subsection{Data Collection}

\label{sec:data_collection}
\paragraph{\textbf{Sample Image Set}}
Over 10,000 scientific images are collected from publicly available datasets, categorized into three groups:  
(1) \textit{general scientific understanding datasets},  
(2) \textit{domain-specific expert datasets} from STEM fields, and  
(3) \textit{AI-generated images} created by text-to-image models using scientific prompts.
Only raw images are extracted, and near-duplicates are removed via image hashing for datasets exceeding 10,000 samples, ensuring a clean image pool.
Each image is assigned to one of seven functional categories to characterize its intended scientific role.
Finally, balanced sampling is performed by selecting up to 2,000 images per category, resulting in an image set of 11,515 unique scientific images. Detailed information is provided in Appendix~\ref{app:data_sources}.

\paragraph{\textbf{MCQ Set Generation of SIQA-U}}

\begin{figure}[t]
    \centering
    \includegraphics[width=1\linewidth]{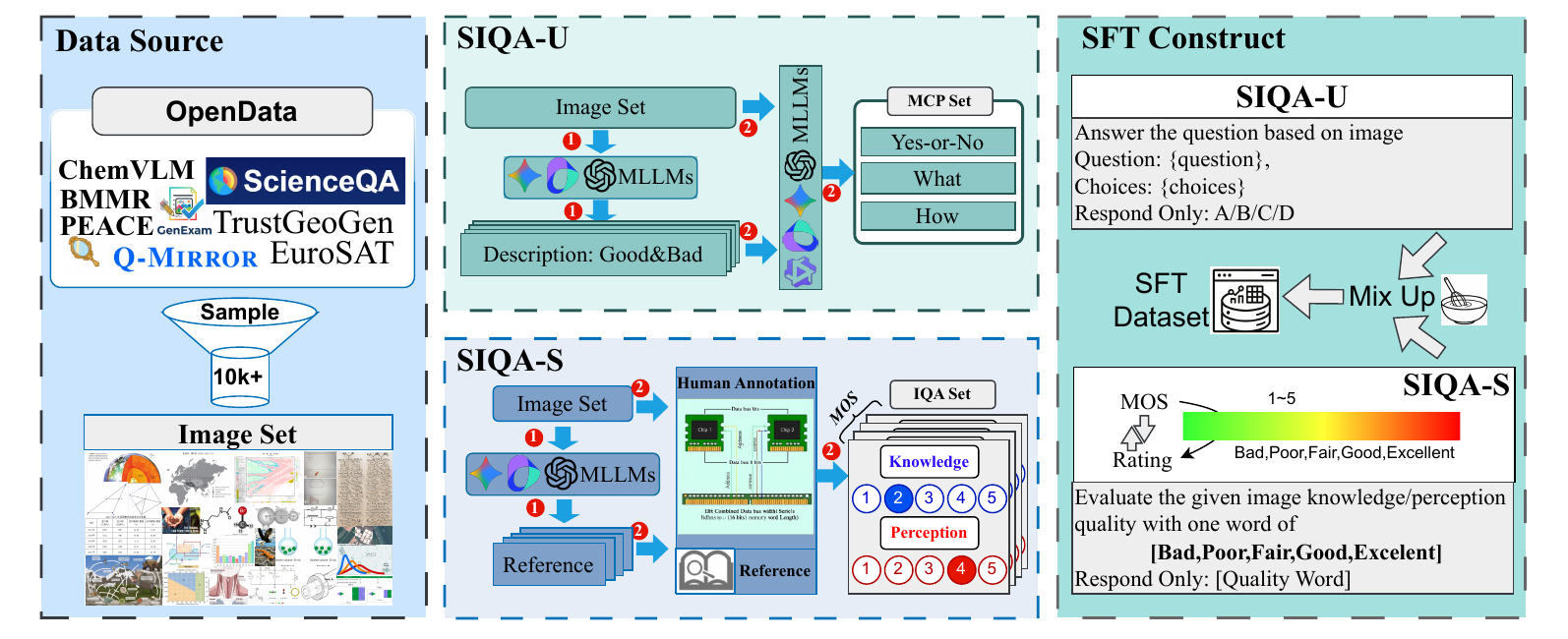}
    \caption{The construction pipeline of the SIQA Challenge, detailing image sourcing, MCQ generation, and annotation strategies. By combining MLLM-driven generation with rigorous human expert validation, the pipeline ensures high-quality data construction, yielding both a large-scale Training Set for SFT and a strictly isolated Benchmark for robust performance evaluation.}
    \label{fig:data_collection}
    \vspace{-1.5em}
\end{figure}

As illustrated in Fig.~\ref{fig:data_collection} (SIQA-U), a candidate pool of 180k+ multiple-choice questions (MCQs) is generated from 11,515 images to support downstream benchmark construction. 
To ensure semantic diversity and alignment with the target evaluation dimensions, the generation process is decomposed into two stages:
(1) For each of the four SIQA dimensions, MLLMs are prompted to produce descriptions of scientific images that explicitly contrast ``good'' and ``bad'' qualities along that dimension.
(2)  By iterating over all four dimensions and three question types, 3–6 candidate questions are generated per image, guided by the combination of dimension-specific descriptions with the target question type.
These pseudo answers serve as provisional ground truth and are not verified by model responses at this stage.

\paragraph{\textbf{Benchmark Split}}

The SIQA Challenge is partitioned into a \textbf{TrainSet} for model development and a strictly isolated \textbf{Benchmark} for final evaluation. 
To construct the Benchmark, 20k multiple-choice questions (MCQs) are sampled from the pool of 180k+ MCQs, balancing challenge difficulty with computational cost.
To estimate question difficulty, MLLMs (distinct from those used in generation) are prompted to answer these MCQs. 
Their predictions are compared against the pseudo answers to compute correctness statistics.
Questions are selected to approximate an incorrect-to-correct response ratio of 8:2, prioritizing questions with lower model accuracy to enhance the discriminative capacity of the benchmark.
To avoid redundancy, an image-level deduplication step is applied to the selected MCQ-image pairs. All images selected for the benchmark are strictly isolated and reserved exclusively for evaluation. 
Consequently, any other MCQs associated with these specific images are discarded and excluded from the trainset.
The final benchmark results in 2,240 unique image–MCQ pairs.
Based on this selection, the SIQA Challenge is split without data leakage:
\begin{itemize}
    \vspace{-0.5em}
    \item \textbf{SIQA-U Benchmark}: 2,240 images with corresponding MCQs,
    \item \textbf{SIQA-U TrainSet}: remaining images with 106k+ MCQs,
    \item \textbf{SIQA-S Benchmark}: 2,100 images of 2,240 benchmark images with expert annotations of perception and knowledge,
    \item \textbf{SIQA-S TrainSet}: 8,400 images of trainset images for further annotation.
    \vspace{-0.7em}
\end{itemize}

\paragraph{\textbf{SIQA-U Benchmark Refining}}
To mitigate potential response biases and ensure fair evaluation, targeted refinements are applied to the benchmark questions based on their type:
 \textit{Yes/No}: To prevent MLLMs from exploiting linguistic priors (e.g., a tendency to favor ``Yes''), the options are rewritten using negated or logically reversed phrasing (e.g., changing “Is this diagram correct?” to “Does this diagram contain errors?”), thereby balancing the semantic polarity of the choices.
 \textit{What}: To eliminate positional bias, the order of the four answer choices is randomly shuffled  during finalization, ensuring no systematic advantage for any option position (A–D).
 \textit{How}: Since the options correspond to an ordinal quality scale (Bad to Excellent), shuffling would disrupt interpretability. Thus, the natural order is retained without any rebalancing.
Pilot experiments (see Appendix~\ref{app:siqa_u_construction}) demonstrate that these refinements eliminate significant biases in the final benchmark.

\begin{figure}[t]
    \centering
    \includegraphics[width=0.7\linewidth]{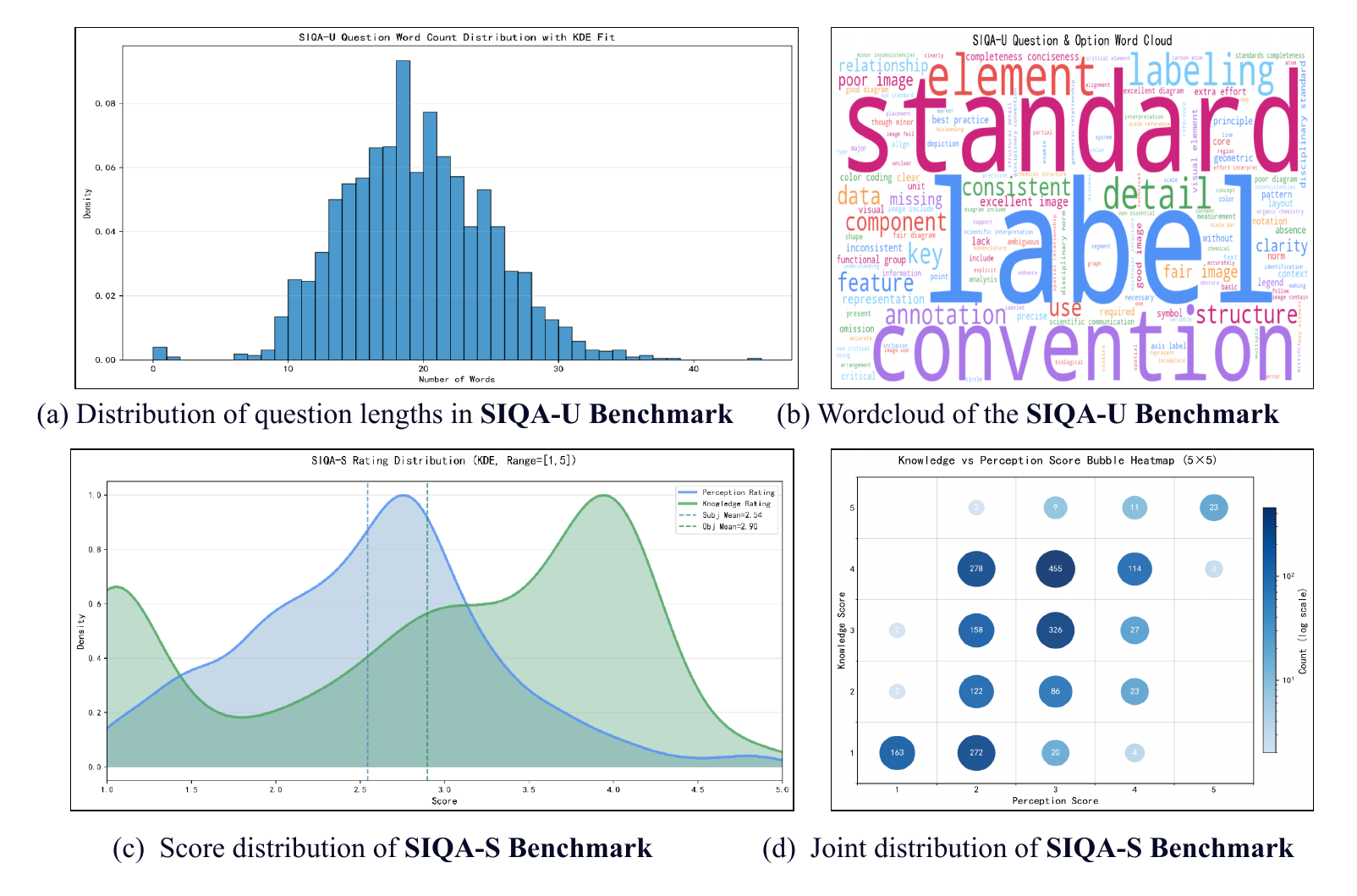}
    \caption{Statistical properties of the SIQA benchmark. 
    (a) Question length distribution in SIQA-U,
    (b) Word cloud visualizing domain-specific terminology in SIQA-U, 
    (c) Score distributions for Perceptual Quality and Knowledge-based Quality in SIQA-S, 
    (d) Joint distribution of Perceptual Quality and Knowledge-based Quality scores in SIQA-S, revealing their correlation structure.}
    \label{fig:benchmark_distribution}
    \vspace{-1.5em}
\end{figure}

\subsection{Expert Annotation}
\label{Sec:annotation}

\paragraph{\textbf{Annotation Teams}}
A team of 17 annotators is assembled, each selected based on academic credentials and domain expertise. All annotators are either professionals employed in STEM-related fields or graduate and undergraduate students with substantial prior engagement with scientific literature and visualizations, ensuring a solid foundation for evaluating both perceptual clarity and scientific content.

\paragraph{\textbf{Benchmark Annotation}}
For SIQA-U, final answers are determined strictly by expert consensus. Unlike other subsets, these labels are generated entirely through human expert deliberation, bearing no reliance on model outputs during the benchmark construction. 
For \textbf{SIQA-S}, annotators rate each image on a five-point scale across perception and knowledge dimensions (see Appendix~\ref{app:siqa_s_annotation}). 
To ensure reliable ratings, distinct rubrics are provided for each dimension, and all annotators are required to pass a calibration test on a representative set of 30 images. 
Using a consensus score established by senior annotators, only those achieving a Spearman Rank Correlation Coefficient (SRCC) above 0.8 with a reference consensus score are permitted to proceed.
Each annotation session is limited to approximately 60 minutes, after which annotators are required to rest for at least 30 minutes.

For \textbf{SIQA-U}, a semi-automatic yet verified annotation strategy is adopted. Multiple MLLMs independently answer each MCQ, and majority voting determines the final label. A randomly sampled subset is audited by experts and batches exhibiting systematic deviations are re-annotated through the same procedure.
For \textbf{SIQA-S}, all MOS ratings are provided exclusively by the top-performing annotators who demonstrate the highest SRCC on the SIQA-S Benchmark, thereby ensuring consistent score distributions between the Benchmark and the TrainSet.

\subsection{Data Analysis}
\label{Sec:anaysis}

Question lengths are centered around 20 tokens (15–25 range), as shown in Fig.~\ref{fig:benchmark_distribution}(a), indicating that the questions extend beyond trivial prompts and require non-trivial scientific reasoning.
Fig.~\ref{fig:benchmark_distribution}(b) shows that frequently occurring terms align with the four SIQA dimensions: words such as ``consistent'' and ``missing'' are associated with \textit{Scientific Validity} and \textit{Completeness}, while ``clarity'' and ``convention'' correspond to \textit{Cognitive Clarity} and \textit{Disciplinary Conformity}, reflecting semantic coverage of the proposed framework.

For SIQA-S, the MOS distributions in Fig.~\ref{fig:benchmark_distribution}(c) show that perception scores exhibit a unimodal pattern with decreasing frequency at higher rating levels.
Knowledge scores display a multi-modal distribution, including a concentration at lower score levels, consistent with the inclusion of images with limited scientific grounding (e.g., recognition-oriented visuals) during data collection.
Finally, the joint distribution in Fig.~\ref{fig:benchmark_distribution}(d) reveals a statistically significant yet moderate association between perception and knowledge (SRCC = 0.587, PLCC = 0.651). High performance along one axis does not necessarily imply high performance along the other; for instance, images rated at perception level 2 span nearly the full range of knowledge scores.

Together, these observations suggest that perception and knowledge capture related but partially distinct facets of scientific image quality.

\subsection{Data Preparation for SFT}
\label{Sec:model SFT}

To enable supervised fine-tuning (SFT) of MLLMs, the dataset is reformulated into an instruction-following format. For \textbf{SIQA-U}, after defining a standardized system prompt, the final input combines the MCQ without answer into a unified prompt. For \textbf{SIQA-S}, the continuous MOS is converted into five discrete quality ratings \{\texttt{Bad}, \texttt{Poor}, \texttt{Fair}, \texttt{Good}, and \texttt{Excellent}\} via uniform binning over the interval $[m, M]$, as defined by:
\begin{align}
    R(s) = r_i \text{ if }m+\frac{i-1}{5}(M-m) \leq s < m+\frac{i}{5}(M-m), \\
    \{r_i\}_{i=1}^{5} = \{\texttt{Bad}, \texttt{Poor}, \texttt{Fair}, \texttt{Good}, \texttt{Excellent}\}.
\end{align}
The final prompt is constructed by adding a standardized system instruction to the task input.
Detailed prompt templates are provided in the Appendix~\ref{app:exp_setup}. All models are trained using only the standard language modeling loss to ensure a fair comparison of architecture and scale effects.
\begin{align}
    \mathcal{L} = -\sum_t \log P(x_t \mid x_{<t}; \theta)
\end{align}

\section{Experiments}

\subsection{Evaluation Setup}

\paragraph{\textbf{SIQA-U:}} Evaluation only requires the model’s final answer, therefore we evaluate a broad set of mainstream MLLMs, including
\textit{O3~\cite{o3}}
\textit{GPT-5~\cite{gpt5}}
\textit{GPT-4o~\cite{gpt4o}}, 
\textit{GPT-3.5-turbo~\cite{gpt35}},
\textit{Gemini-2.5-pro~\cite{gemini25}}, 
\textit{Claude-sonnet-4.5~\cite{claude45}}
\textit{Doubao-1.5-Vision-pro~\cite{doubao15}}, 
\textit{GLM-4.6v~\cite{GLM46v}}
\textit{DeepSeek-VL2~\cite{dsvl2}}, 
\textit{LLaMA-3.2-90B-Vision~\cite{llama32}}.
\textit{Qwen3-VL-235B~\cite{Qwen3-VL}}, and
\textit{InternVL3.5-241B~\cite{wang2025internvl3}}.
On SIQA-U, evaluation is based solely on the final option (A–D). While models are prompted to output a single character, we extract the predicted answer whenever it is clearly identifiable, regardless of extra text. Responses failing this criterion are marked incorrect.

\paragraph{\textbf{SIQA-S:}} Evaluation requires both the predicted quality token and its corresponding logits, so our evaluation is limited to models with publicly available weights or logprob support. We therefore evaluate the following MLLMs on SIQA-S: \textit{GPT-4o}, \textit{GPT-3.5-turbo} and \textit{Doubao-1.5-Vision-pro} (via API with logprobs), as well as \textit{Qwen3-VL-235B}, \textit{InternVL3.5-241B} and \textit{Llama-3.2-90B-Vision}.
We extract the logits corresponding to all five rating levels from the model’s output and compute the predicted score as:
\begin{align}
    S_{pred} = \sum_{i=1}^5p_{l_i}G(l_i)=\sum_{i=1}^5i\times\frac{e^{\mathcal{X}_{l_i}}}{\sum_{j=1}^5e^{\mathcal{X}_{l_j}}}
    \label{Formula:ratings}
\end{align}
Where the $l_i\in \{Bad, Poor, Fair, Good, Excellent\}$ denotes the $i-$th quality level, $G(l_i)=i$ is its numerical ratings and $\mathcal{X}_{l_i}$ is the model's logits of token representing level $l_i$. 
These predicted scores are then used to compute the SRCC and Pearson Linear Correlation Coefficient (PLCC) for both the perception and knowledge dimensions.

\subsection{Findings of Evaluation}
\begin{table}[t]
  \centering
  \caption{Performance of MLLMs on the SIQA-U benchmark. Results are reported as accuracy across four scientific quality dimensions: Scientific Completeness (Comp.), Scientific Validity (Val.), Disciplinary Conformity (Conf.), and Cognitive Clarity (Clar.), and three question types (Yes/No, What, How). \textbf{Bold}: best; \underline{Underline}: second-best.}

  \label{tab:SIQA-U}
  \scriptsize
  \setlength{\tabcolsep}{3pt}
\resizebox{\linewidth}{!}{
    \begin{tabular}{l | cccc | ccc | c}
      \hline\hline
        \textbf{Model} 
        & Comp. 
        & Val. 
        & Conf. 
        & Clar. 
        & Yes/No 
        & What 
        & How 
        & Overall \\
      \hline
      \multicolumn{9}{l}{\textit{Closed-Source}} \\
      \cdashline{1-9}
      O3~\cite{o3}                                  & 0.424 & 0.434 & 0.465 & 0.434 & 0.417 & 0.648 & 0.255 & 0.435 \\
      GPT-5~\cite{gpt5}                             & 0.441 & 0.455 & 0.479 & 0.455 & 0.448 & 0.677 & 0.243 & 0.450 \\
      GPT-4o~\cite{gpt4o}                           & 0.427 & 0.421 & 0.477 & 0.421 & 0.426 & 0.612 & 0.283 & 0.436 \\
      GPT-3.5-turbo~\cite{gpt35}                    & 0.399 & 0.460 & 0.459 & 0.460 & 0.444 & 0.595 & 0.271 & 0.432 \\
      Gemini-2.5-Pro~\cite{gemini25}                & 0.448 & 0.482 & 0.479 & 0.482 & 0.462 & 0.700 & 0.256 & 0.467 \\
      Claude-sonnet-4.5~\cite{claude45}             & 0.430 & \underline{0.507} & \underline{0.508} & \underline{0.507} & 0.435 & \underline{0.716} & \textbf{0.300} & \underline{0.478} \\
      Doubao-1.5-Vision-Pro~\cite{doubao15}  & 0.437 & 0.429 & 0.486 & 0.429 & 0.478 & 0.642 & 0.241 & 0.449 \\
      \hline
      \multicolumn{9}{l}{\textit{Open-Source}} \\
      \cdashline{1-9}
      GLM-4.6v~\cite{GLM46v}                        & 0.458 & 0.506 & 0.443 & 0.459 & 0.448 & 0.676 & 0.285 & 0.468 \\
      DeepSeek-VL2~\cite{dsvl2}                     & 0.330 & 0.421 & 0.433 & 0.421 & 0.396 & 0.503 & 0.273 & 0.388 \\
      LLaMA-3.2-90B-Vision~\cite{llama32}           & 0.468 & 0.444 & 0.472 & 0.444 & \underline{0.483} & 0.620 & 0.286 & 0.459 \\
      InternVL3.5-241B~\cite{wang2025internvl3}& \underline{0.479} & 0.469 & 0.461 & 0.484 & 0.465 & 0.686 & 0.271 & 0.473 \\
      Qwen3-VL-235B~\cite{Qwen3-VL}   & 0.459 & 0.451 & 0.485 & 0.451 & 0.459 & 0.660 & 0.270 & 0.458 \\
      \hline
      \textbf{SIQA-Judger} (trained)                & \textbf{0.605} & \textbf{0.546} & \textbf{0.525} & \textbf{0.571} & \textbf{0.574} & \textbf{0.830} & \underline{0.290} & \textbf{0.563}\\
      \cdashline{1-9}
      Random Guess                                  & 0.365 & 0.336  & 0.339 & 0.361 & 0.500 & 0.250  & 0.250  & 0.350 \\
      \hline\hline
    \end{tabular}%
}
\vspace{-1.5em}
\end{table}


\begin{figure}[t]
    \centering
        \includegraphics[width=\linewidth]{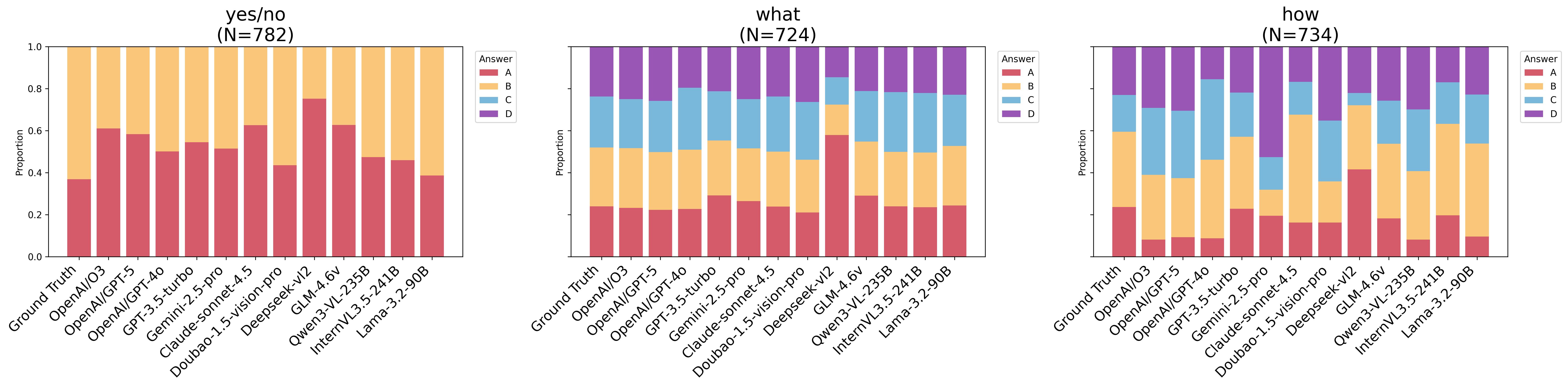}
        \caption{Answer distributions of different MLLMs on SIQA-U.} 
        \label{fig:answer distrubtion}
        \vspace{-1.5em}
\end{figure}

\paragraph{\textbf{Finding of SIQA-U: Current MLLMs can accurately describe the visual content of scientific figures but fail to validate their scientific correctness or explain their quality.}} 
As shown in Table~\ref{tab:SIQA-U}, models achieve 76.4\% accuracy on \textit{What}-type questions (e.g.,``What is shown in the figure?''), significantly above the 25\% random baseline. 
In contrast, they score only 48.3\% on binary \textit{Yes/No} verification, slightly below the 50\% random chance, and 30.0\% on \textit{How}-type justification, barely above random guessing. 
This poor performance stems from systematic response biases rather than informed reasoning, as shown in Fig.~\ref{fig:answer distrubtion}.
While most models favor option ``A'' on \textit{Yes/No} questions, DeepSeek-VL2 exhibits the strongest bias, which continues to dominate even in \textit{What}-type questions.
On \textit{How}-type questions, requiring evaluation of scientific quality, while most models exhibit a conservative bias by rarely assigning the ``A. Excellent'' rating, DeepSeek-VL2 shows the opposite trend, overwhelmingly favoring Option A.
In contrast, Gemini-2.5-Pro predominantly selects option D, revealing a highly skewed response distribution. 
This suggests that although MLLMs possess basic perceptual competence, they lack the ability to apply domain knowledge to judge scientific validity.
We hypothesize that this gap stems from standard pretraining: MLLMs are typically optimized with cross-entropy loss on image–caption pairs, which lack supervisory signal for judging the factual correctness or scientific quality of the image content itself.

\begin{table}[t]
\centering
\caption{
  Performance of various models on the SIQA-S benchmark.
  Correlation metrics (SRCC\&PLCC) are reported for Perception and Knowledge and the average as SIQA-S. \textbf{Bold}: best; \underline{Underline}: second-best.
}

\label{tab:scoring}
\scriptsize
\renewcommand\arraystretch{1.20}
\setlength{\tabcolsep}{7pt} 
\resizebox{\linewidth}{!}{
\begin{tabular}{l | c c | c c | c c}
\hline\hline
\textbf{Dim}
& \multicolumn{2}{c|}{Perception} 
& \multicolumn{2}{c|}{Knowledge} 
& \multicolumn{2}{c}{Overall} \\
\hline
\textbf{Model} & SRCC & PLCC & SRCC & PLCC & SRCC & PLCC \\
\hline
\multicolumn{7}{l}{\textit{Closed-Source MLLM (zero-shot)}} \\
\cdashline{1-7}
GPT-4o~\cite{gpt4o}                     & 0.694&0.699 & 0.836& 0.850 & 0.765 & 0.775 \\
GPT-3.5-turbo~\cite{gpt35}              & 0.480&0.425 & 0.699&0.665 & 0.589&0.545 \\
Doubao-1.5-Vision-Pro~\cite{doubao15}   & 0.596 & 0.669 & 0.799 & 0.845 & 0.698 & 0.757 \\ 
\hline
\multicolumn{7}{l}{\textit{Open-Source MLLM (zero-shot)}} \\  %
\cdashline{1-7}
Llama-3.2-90B-Vision~\cite{llama32}         & 0.568 & 0.611 & 0.482 & 0.580 & 0.525 & 0.596 \\
InternVL3.5-241B~\cite{wang2025internvl3}   & 0.600 & 0.638 & 0.835 & 0.835 & 0.717 & 0.736 \\
Qwen3-VL-235B~\cite{Qwen3-VL}               & 0.629 & 0.678 & 0.862 & 0.831 & 0.746 & 0.755 \\
\hline
\multicolumn{7}{l}{\textit{Natural IQA}} \\  %
\cdashline{1-7}
NIQE~\cite{NIQE} (zero-shot)       & 0.345 & 0.235 & 0.447 & 0.410 & 0.396 & 0.322\\
Q-Align~\cite{Q-Align} (zero-shot) & 0.749 & 0.762 & 0.285 & 0.400 & 0.517 & 0.581 \\
CLIP-IQA~\cite{CLIPIQA} (zero-shot)& 0.496 & 0.520 & 0.362 & 0.435 & 0.429 & 0.478\\
CLIP-IQA+~\cite{CLIPIQA} (trained) & 0.724 & 0.676 & 0.862 & 0.801 & 0.793 & 0.741\\
HyperIQA~\cite{CNN-IQA1} (trained) & \underline{0.773} & \underline{0.783} & \underline{0.897} & \underline{0.895} & \underline{0.835} & \underline{0.839}\\ 
\hline
\textbf{SIQA-Judger} (trained)  & \textbf{0.857}&\textbf{0.881} & \textbf{0.915}&\textbf{0.937} & \textbf{0.886}&\textbf{0.909} \\
\hline\hline
\end{tabular}
}
\vspace{-1.5em}
\end{table}

\paragraph{\textbf{Finding of SIQA-S: MLLMs align more closely with humans on knowledge  than on perceptual.}} 
As shown in Table~\ref{tab:scoring}, MLLMs exhibit strong correlation with human raters on the \textit{knowledge} dimension (SRCC 0.7–0.8), which assesses the factual correctness of scientific content in images. 
In stark contrast, their alignment on the perception dimension remains notably weaker (SRCC 0.5–0.6).  
But Q-Align, trained solely on generic visual attributes, outperforms all MLLMs on perception (SRCC 0.749) but collapses on knowledge (SRCC 0.285), as it lacks the capacity to reason about scientific truth. Similarly, zero-shot CLIP-IQA underperforms all MLLMs on both dimensions, particularly on knowledge (SRCC < 0.3), as its fixed semantic matching mechanism cannot dynamically adapt prompts to the image content for assessment.
More importantly, even after task-specific training, traditional methods remain inferior to MLLMs fine-tuned on our data. This persistent gap demonstrates that MLLMs are better suited for SIQA thanks to their pre-trained world knowledge and flexible reasoning capabilities.
The result implies that MLLMs have internalized domain-specific factual knowledge to a degree that closely mirrors human judgment, effectively functioning as reliable validators of scientific plausibility through large-scale pretraining on scientific corpora. 
However, they still struggle to evaluate the perceptual quality of a scientific figure , which likely stems from the fact that training data extensively describes what is shown but rarely judges how well it is presented.

\begin{figure}[t]
    \centering
    \includegraphics[width=0.4\linewidth]{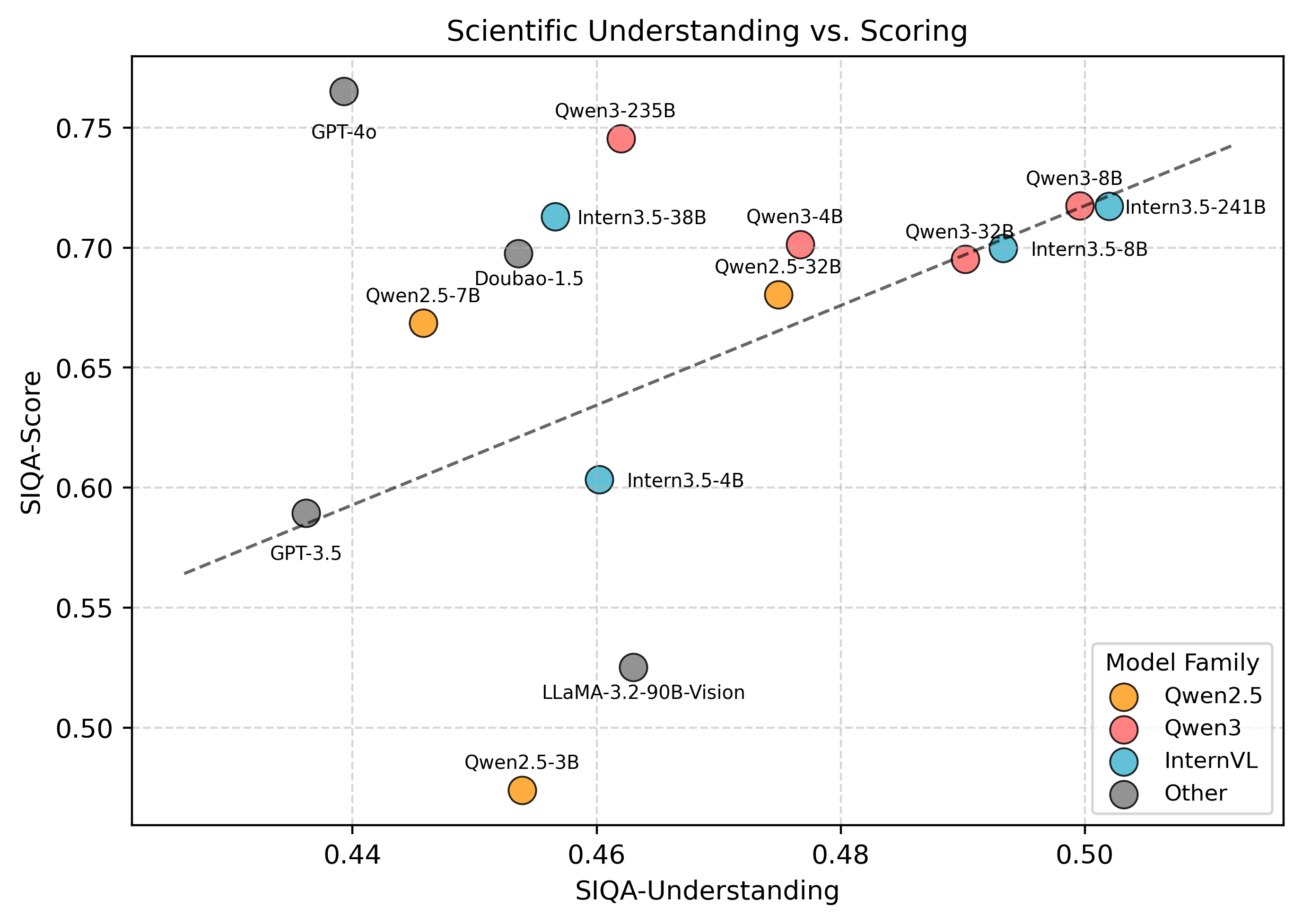}
    \includegraphics[width=0.4\linewidth]{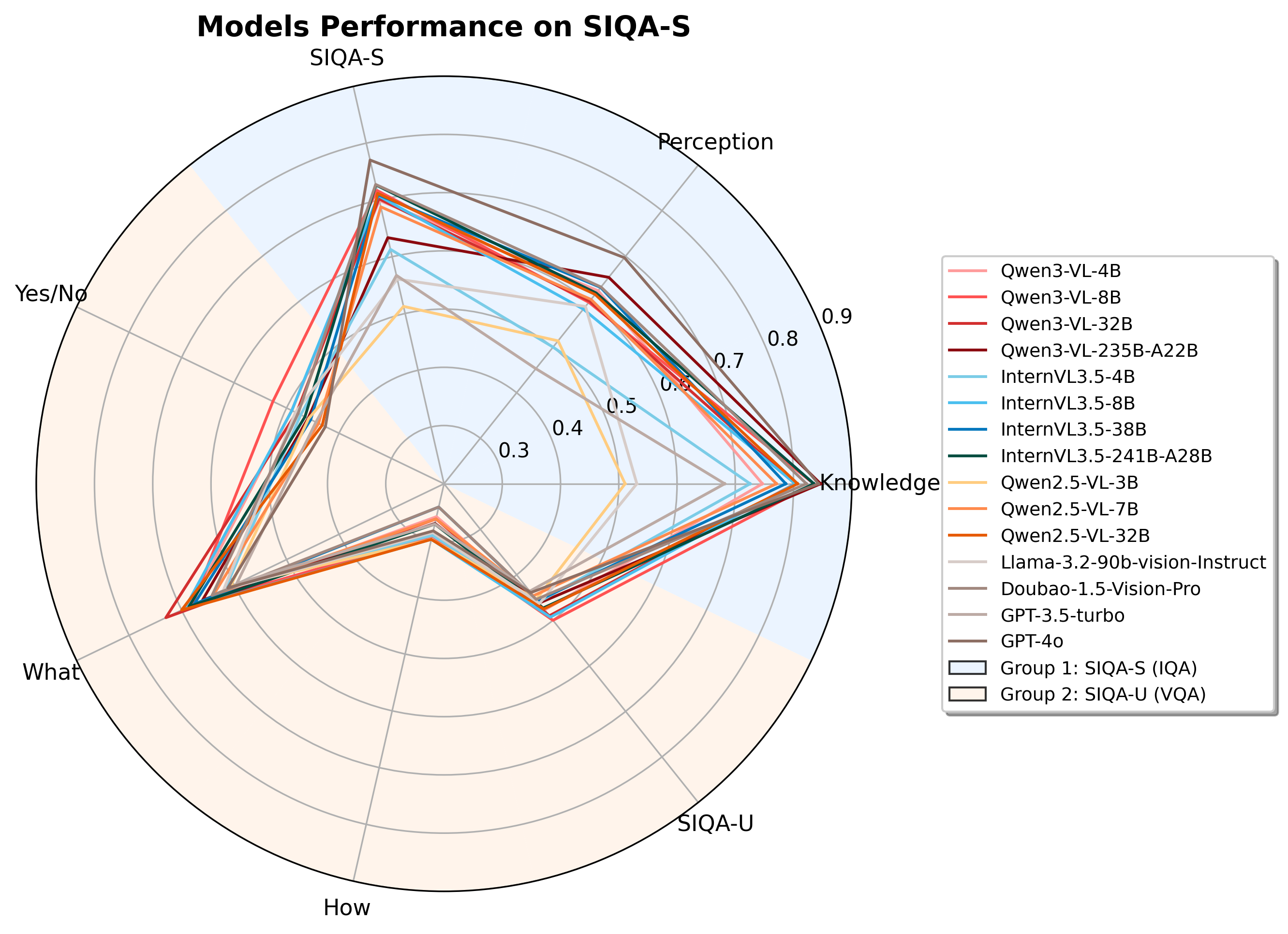}
    \caption{Evaluation of SIQA across MLLMs. Points are color-coded by model family. Results show a positive correlation between SIQA-U and SIQA-S.}
    \label{fig:U-S}
    \vspace{-1.5em}
\end{figure}

\paragraph{\textbf{Finding from SIQA Evaluation: Scoring accuracy generally improves with scientific understanding.}}
Fig.~\ref{fig:U-S} shows a positive correlation between SIQA-S and SIQA-U across various MLLMs. Models with higher accuracy on SIQA-U tend to align better with human judgment in SIQA-S, indicating that better understanding facilitates more accurate scoring.
However, some models deviate from this trend. Notably, GPT-4o and Qwen3-A22B achieve high SIQA-S scores despite moderate SIQA-U performance, while smaller models like Qwen2.5-3B underperform in both metrics, likely due to limited reasoning capacity.
Interestingly, LLaMA-3.2-90B-Vision, despite its large scale, performs poorly in both SIQA-U and SIQA-S, suggesting that model size alone does not guarantee scientific reasoning ability.
In contrast, Qwen3 and InternVL consistently outperform Qwen2.5, highlighting the importance of architectural design and training data quality over size alone.
While SIQA-U generally supports SIQA-S, it is not always a perfect predictor. Some models may achieve high scores through pattern memorization rather than true scientific reasoning, underscoring the need for AI systems that prioritize genuine understanding over surface-level alignment with human judgments.

\subsection{Impact of Supervised Fine-tuning}

We perform full-parameter supervised fine-tuning on a diverse set of models spanning various scales, including the Qwen2.5-VL, Qwen3-VL, and InternVL3.5 families (ranging from 3B to 38B).
As shown in Table~\ref{tab:fine-tune}, fine-tuning drives all models to high alignment, with SIQA-S correlations converging to SRCC around 0.90, regardless of model size or architecture.
However, the improvement in comprehension, measured by SIQA-U accuracy, is modest, rising from 44.4\% to 56.6\%. This discrepancy arises because the scoring task is structurally constrained. Outputs follow rigid templates (e.g., ``Knowledge: Excellent'' or ``Perception: Fair''), allowing models to learn shallow mappings between inputs and scores through label fitting. In contrast, SIQA-U evaluates factual understanding with closed-form questions that require true interpretation of scientific content. 
Fine-tuning focuses on achieving surface plausibility rather than conceptual grounding, causing models to produce calibrated scores that mask their deficits in scientific reasoning.

\begin{table}[t]
\centering
\caption{
Effect of model architecture and scale on fine-tuning to \textbf{SIQA-Judger}. 
We report SIQA-U accuracy (ACC) and SIQA-S correlation metrics (average of SRCC and PLCC) before (Pre) and after (Post) fine-tuning on the SIQA trainset. 
}
\label{tab:fine-tune}

\setlength{\tabcolsep}{3pt}  

\resizebox{\columnwidth}{!}{%
\begin{tabular}{l | c c   c c | c c  c c | c c  c c}
\hline\hline
\textbf{Dimension} & \multicolumn{4}{c|}{Perception}   & \multicolumn{4}{c|}{Knowledge} & \multicolumn{4}{c}{Overall}\\
\hline
\multirow{2}{*}{\textbf{Model}} &  \multicolumn{2}{c}{SIQA-U} & \multicolumn{2}{c|}{SIQA-S} & \multicolumn{2}{c}{SIQA-U} & \multicolumn{2}{c|}{SIQA-S} & \multicolumn{2}{c}{SIQA-U} & \multicolumn{2}{c}{SIQA-S}\\
\cdashline{2-13}
 & Pre & Post & Pre & Post & Pre & Post & Pre & Post & Pre & Post & Pre & Post\\
\hline
\multicolumn{13}{l}{\textit{Qwen2.5-VL}~\cite{Qwen3-VL}} \\
\cdashline{1-13}
Qwen2.5-VL-32B & 0.469 & 0.545 & 0.616 & 0.899 & 0.479 & 0.542 & 0.807 & 0.928 & 0.474 & 0.543 & 0.712 & 0.913 \\
Qwen2.5-VL-7B  & 0.442 & 0.543 & 0.605 & 0.896 & 0.446 & 0.557 & 0.771 & 0.927 & 0.444 & 0.551 & 0.688 & 0.911 \\
Qwen2.5-VL-3B  & 0.441 & 0.524 & 0.514 & 0.891 & 0.467 & 0.560 & 0.511 & 0.924 & 0.455 & 0.543 & 0.513 & 0.907 \\
\hline
\multicolumn{13}{l}{\textit{Qwen3-VL}~\cite{Qwen3-VL}} \\
\cdashline{1-13}
Qwen3-VL-32B   & 0.478 & 0.550 & 0.601 & 0.907 & 0.497 & 0.562 & 0.803 & 0.928 & 0.488 & 0.556 & 0.702 & 0.918 \\
Qwen3-VL-8B    & 0.509 & 0.554 & 0.599 & 0.900 & 0.491 & 0.547 & 0.835 & 0.927 & 0.499 & 0.550 & 0.717 & 0.914 \\
Qwen3-VL-4B    & 0.473 & 0.547 & 0.624 & 0.911 & 0.479 & 0.549 & 0.747 & 0.924 & 0.476 & 0.548 & 0.699 & 0.916 \\
\hline
\multicolumn{13}{l}{\textit{InternVL3.5}~\cite{wang2025internvl3}} \\
\cdashline{1-13}
InternVL3.5-38B & 0.438 & 0.569 & 0.631 & 0.884 & 0.473 & 0.565 & 0.787 & 0.933 & 0.457 & 0.566 & 0.709 & 0.908 \\
InternVL3.5-8B  & 0.493 & 0.547 & 0.584 & 0.891 & 0.490 & 0.557 & 0.803 & 0.931 & 0.491 & 0.552 & 0.708 & 0.911 \\
InternVL3.5-4B  & 0.451 & 0.548 & 0.500 & 0.869 & 0.467 & 0.576 & 0.726 & 0.926 & 0.459 & 0.563 & 0.613 & 0.897 \\
\hline
\hline
\end{tabular}%
}
\vspace{-1.5em}
\end{table}%

\section{Conclusion}

We introduce \textsc{SIQA}, a framework for scientific image quality assessment, defining quality across two dimensions: Knowledge (Scientific Validity and Completeness) and Perception (Cognitive Clarity and Disciplinary Conformity). We also present two evaluation protocols: SIQA-U and SIQA-S, and release the first benchmark with fine-grained human annotations for scientific images. 
Our experiments reveal a critical gap: fine-tuned models achieve high scoring fidelity (SRCC $\approx$ 0.91) but show only marginal improvements in genuine understanding (accuracy increases from 44.4\% to 56.6\%). This highlights that high scores often result from surface-level pattern matching rather than true scientific reasoning.
\textsc{SIQA} provides the necessary tools to distinguish performative fluency from authentic competence, setting the stage for trustworthy AI in scientific practice, where both perceptual clarity and knowledge correctness are essential.

\bibliographystyle{splncs04}
\bibliography{main}

@String(CVPR  = {IEEE Conf. Comput. Vis. Pattern Recog.})

@String(NeurIPS = {Adv. Neural Inform. Process. Syst.})

@String(AAAI  = {AAAI})

@String(CVPR  = {CVPR})

@String(NeurIPS = {NeurIPS})

@inproceedings{MPS,
            title={Learning Multi-dimensional Human Preference for Text-to-Image Generation},
            author={Zhang, Sixian and Wang, Bohan and Wu, Junqiang and Li, Yan and Gao, Tingting and Zhang, Di and Wang, Zhongyuan},
            booktitle={Proceedings of the IEEE/CVF Conference on Computer Vision and Pattern Recognition},
            pages={8018--8027},
            year={2024}
          }

@article{Brecher+2008+277+410,
title = {Graphical representation standards for chemical structure diagrams (IUPAC Recommendations 2008)},
author = {Jonathan Brecher},
pages = {277--410},
volume = {80},
number = {2},
journal = {Pure and Applied Chemistry},
doi = {doi:10.1351/pac200880020277},
year = {2008},
lastchecked = {2026-01-29}
}

@book{national2019reproducibility,
  author    = {{National Academies of Sciences, Engineering, and Medicine}},
  title     = {Reproducibility and Replicability in Science},
  year      = {2019},
  month     = may,
  day       = {7},
  publisher = {National Academies Press (US)},
  address   = {Washington, DC},
  isbn      = {978-0-309-48616-3},
  doi       = {10.17226/25303},
  url       = {https://nap.nationalacademies.org/catalog/25303/reproducibility-and-replicability-in-science},
  pmid      = {31596559},
  note      = {Report by the Committee on Reproducibility and Replicability in Science}
}

@misc{liu2023improvedllava,
          author={Liu, Haotian and Li, Chunyuan and Li, Yuheng and Lee, Yong Jae},
          title={Improved Baselines with Visual Instruction Tuning}, 
          publisher={arXiv:2310.03744},
          year={2023},
  }

@inproceedings{liu2023llava,
    author      = {Liu, Haotian and Li, Chunyuan and Wu, Qingyang and Lee, Yong Jae},
    title       = {Visual Instruction Tuning},
    booktitle   = {NeurIPS},
    year        = {2023}
  }

@article{zhao2024chemdfm,
  title={ChemDFM: a large language foundation model for chemistry},
  author={Zhao, Zihan and Ma, Da and Chen, Lu and Sun, Liangtai and Li, Zihao and Xia, Yi and Chen, Bo and Xu, Hongshen and Zhu, Zichen and Zhu, Su and others},
  journal={arXiv preprint arXiv:2401.14818},
  year={2024}
}

@inproceedings{cao2025instructmol,
  title={Instructmol: Multi-modal integration for building a versatile and reliable molecular assistant in drug discovery},
  author={Cao, He and Liu, Zijing and Lu, Xingyu and Yao, Yuan and Li, Yu},
  booktitle={Proceedings of the 31st International Conference on Computational Linguistics},
  pages={354--379},
  year={2025}
}

@article{lv2023kosmos,
  title={Kosmos-2.5: A multimodal literate model},
  author={Lv, Tengchao and Huang, Yupan and Chen, Jingye and Zhao, Yuzhong and Jia, Yilin and Cui, Lei and Ma, Shuming and Chang, Yaoyao and Huang, Shaohan and Wang, Wenhui and others},
  journal={arXiv preprint arXiv:2309.11419},
  year={2023}
}

@article{lala2023paperqa,
  title={Paperqa: Retrieval-augmented generative agent for scientific research},
  author={L{\'a}la, Jakub and O'Donoghue, Odhran and Shtedritski, Aleksandar and Cox, Sam and Rodriques, Samuel G and White, Andrew D},
  journal={arXiv preprint arXiv:2312.07559},
  year={2023}
}

@article{zhang2025large,
  author    = {Zhang, Zicheng and Wang, Junying and Wen, Farong and Guo, Yijin and others},
  title     = {Large Multimodal Models Evaluation: A Survey},
  journal   = {SCIENCE CHINA Information Sciences},
  year      = {2025},
  number  = {12},
  volume    = {68},
  pages     = {221301-221369},
  doi       = {https://doi.org/10.1007/s11432-025-4676-4}
}

@article{Q-Align,
  title={Q-align: Teaching lmms for visual scoring via discrete text-defined levels},
  author={Wu, Haoning and Zhang, Zicheng and Zhang, Weixia and Chen, Chaofeng and Liao, Liang and Li, Chunyi and Gao, Yixuan and Wang, Annan and Zhang, Erli and Sun, Wenxiu and others},
  journal={arXiv preprint arXiv:2312.17090},
  year={2023}
}

@inproceedings{Q-Instruct,
  title={Q-instruct: Improving low-level visual abilities for multi-modality foundation models},
  author={Wu, Haoning and Zhang, Zicheng and Zhang, Erli and Chen, Chaofeng and Liao, Liang and Wang, Annan and Xu, Kaixin and Li, Chunyi and Hou, Jingwen and Zhai, Guangtao and others},
  booktitle={Proceedings of the IEEE/CVF conference on computer vision and pattern recognition},
  pages={25490--25500},
  year={2024}
}

@article{ML-IQA1,
  title={Image quality assessment: from error visibility to structural similarity},
  author={Wang, Zhou and Bovik, Alan C and Sheikh, Hamid R and Simoncelli, Eero P},
  journal={IEEE transactions on image processing},
  volume={13},
  number={4},
  pages={600--612},
  year={2004},
  publisher={IEEE}
}

@article{ML-IQA2,
  title={No-reference image quality assessment in the spatial domain},
  author={Mittal, Anish and Moorthy, Anush Krishna and Bovik, Alan Conrad},
  journal={IEEE Transactions on image processing},
  volume={21},
  number={12},
  pages={4695--4708},
  year={2012},
  publisher={IEEE}
}

@InProceedings{CNN-IQA1,
author = {Su, Shaolin and Yan, Qingsen and Zhu, Yu and Zhang, Cheng and Ge, Xin and Sun, Jinqiu and Zhang, Yanning},
title = {Blindly Assess Image Quality in the Wild Guided by a Self-Adaptive Hyper Network},
booktitle = {Proceedings of the IEEE/CVF Conference on Computer Vision and Pattern Recognition (CVPR)},
month = {June},
year = {2020}
}

@article{CNN-IQA2,
author={Bosse, Sebastian and Maniry, Dominique and Müller, Klaus-Robert and Wiegand, Thomas and Samek, Wojciech},
journal={IEEE Transactions on Image Processing}, 
title={Deep Neural Networks for No-Reference and Full-Reference Image Quality Assessment}, 
year={2018},
volume={27},
number={1},
pages={206-219},
doi={10.1109/TIP.2017.2760518}}

@inproceedings{CLIP,
  title={Learning transferable visual models from natural language supervision},
  author={Radford, Alec and Kim, Jong Wook and Hallacy, Chris and Ramesh, Aditya and Goh, Gabriel and Agarwal, Sandhini and Sastry, Girish and Askell, Amanda and Mishkin, Pamela and Clark, Jack and others},
  booktitle={International conference on machine learning},
  pages={8748--8763},
  year={2021},
  organization={PmLR}
}

@inproceedings{CLIPIQA,
  title={Exploring clip for assessing the look and feel of images},
  author={Wang, Jianyi and Chan, Kelvin CK and Loy, Chen Change},
  booktitle={Proceedings of the AAAI conference on artificial intelligence},
  volume={37},
  pages={2555--2563},
  year={2023}
}

@ARTICLE{NIQE,
  author={Mittal, Anish and Soundararajan, Rajiv and Bovik, Alan C.},
  journal={IEEE Signal Processing Letters}, 
  title={Making a “Completely Blind” Image Quality Analyzer}, 
  year={2013},
  volume={20},
  number={3},
  pages={209-212},
  doi={10.1109/LSP.2012.2227726}}

@article{ghosh2023geneval,
  title={Geneval: An object-focused framework for evaluating text-to-image alignment},
  author={Ghosh, Dhruba and Hajishirzi, Hannaneh and Schmidt, Ludwig},
  journal={Advances in Neural Information Processing Systems},
  volume={36},
  pages={52132--52152},
  year={2023}
}

@article{Qwen3-VL,
      title={Qwen3-VL Technical Report}, 
      author={Shuai Bai and Yuxuan Cai and Ruizhe Chen and Keqin Chen and Xionghui Chen and Zesen Cheng and Lianghao Deng and Wei Ding and Chang Gao and Chunjiang Ge and Wenbin Ge and Zhifang Guo and Qidong Huang and Jie Huang and Fei Huang and Binyuan Hui and Shutong Jiang and Zhaohai Li and Mingsheng Li and Mei Li and Kaixin Li and Zicheng Lin and Junyang Lin and Xuejing Liu and Jiawei Liu and Chenglong Liu and Yang Liu and Dayiheng Liu and Shixuan Liu and Dunjie Lu and Ruilin Luo and Chenxu Lv and Rui Men and Lingchen Meng and Xuancheng Ren and Xingzhang Ren and Sibo Song and Yuchong Sun and Jun Tang and Jianhong Tu and Jianqiang Wan and Peng Wang and Pengfei Wang and Qiuyue Wang and Yuxuan Wang and Tianbao Xie and Yiheng Xu and Haiyang Xu and Jin Xu and Zhibo Yang and Mingkun Yang and Jianxin Yang and An Yang and Bowen Yu and Fei Zhang and Hang Zhang and Xi Zhang and Bo Zheng and Humen Zhong and Jingren Zhou and Fan Zhou and Jing Zhou and Yuanzhi Zhu and Ke Zhu},
	  journal={arXiv preprint arXiv:2511.21631},
      year={2025}
}

@article{Qwen2.5-VL,
  title={Qwen2.5-VL Technical Report},
  author={Bai, Shuai and Chen, Keqin and Liu, Xuejing and Wang, Jialin and Ge, Wenbin and Song, Sibo and Dang, Kai and Wang, Peng and Wang, Shijie and Tang, Jun and Zhong, Humen and Zhu, Yuanzhi and Yang, Mingkun and Li, Zhaohai and Wan, Jianqiang and Wang, Pengfei and Ding, Wei and Fu, Zheren and Xu, Yiheng and Ye, Jiabo and Zhang, Xi and Xie, Tianbao and Cheng, Zesen and Zhang, Hang and Yang, Zhibo and Xu, Haiyang and Lin, Junyang},
  journal={arXiv preprint arXiv:2502.13923},
  year={2025}
}

@article{wang2025internvl3,
    title={InternVL3. 5: Advancing Open-Source Multimodal Models in Versatility, Reasoning, and Efficiency},
    author={Wang, Weiyun and Gao, Zhangwei and Gu, Lixin and Pu, Hengjun and Cui, Long and Wei, Xingguang and Liu, Zhaoyang and Jing, Linglin and Ye, Shenglong and Shao, Jie and others},
    journal={arXiv preprint arXiv:2508.18265},
    year={2025}
}

@misc{gpt5,
  author       = {{OpenAI}},
  title        = {Introducing {GPT-5}},
  howpublished = {\url{https://openai.com/index/introducing-gpt-5/}},
  month        = {August},
  year         = {2025},
  day          = {7},
}

@article{gemini25,
  title={Gemini 2.5: Pushing the frontier with advanced reasoning, multimodality, long context, and next generation agentic capabilities},
  author={Comanici, Gheorghe and Bieber, Eric and Schaekermann, Mike and Pasupat, Ice and Sachdeva, Noveen and Dhillon, Inderjit and Blistein, Marcel and Ram, Ori and Zhang, Dan and Rosen, Evan and others},
  journal={arXiv preprint arXiv:2507.06261},
  year={2025}
}

@misc{o3,
  author       = {{OpenAI}},
  title        = {Introducing {O3} and {O4-mini}},
  howpublished = {\url{https://openai.com/index/introducing-o3-and-o4-mini/}},
  month        = {October},
  year         = {2025},
}

@misc{doubao15,
  author       = {{ByteDance}},
  title        = {Doubao 1.5 Pro},
  howpublished = {\url{https://seed.bytedance.com/en/special/doubao$\_1_5\_$pro}},
  year         = {2025},
}

@misc{claude45,
  author       = {{Anthropic}},
  title        = {Introducing {C}laude 4.5 {S}onnet},
  howpublished = {\url{https://www.anthropic.com/news/claude-sonnet-4-5}},
  year         = {2025},
}

@article{gpt4o,
  title={Gpt-4o system card},
  author={Hurst, Aaron and Lerer, Adam and Goucher, Adam P and Perelman, Adam and Ramesh, Aditya and Clark, Aidan and Ostrow, AJ and Welihinda, Akila and Hayes, Alan and Radford, Alec and others},
  journal={arXiv preprint arXiv:2410.21276},
  year={2024}
}

@misc{gpt35,
  author = {{OpenAI}},
  title = {{GPT-3.5 Turbo}},
  year = {2023},
  howpublished = {\url{https://platform.openai.com/docs/models/gpt-3.5-turbo}},
  note = {Accessed: 2025-12-22}
}

@article{dsvl2,
  title={Deepseek-vl2: Mixture-of-experts vision-language models for advanced multimodal understanding},
  author={Wu, Zhiyu and Chen, Xiaokang and Pan, Zizheng and Liu, Xingchao and Liu, Wen and Dai, Damai and Gao, Huazuo and Ma, Yiyang and Wu, Chengyue and Wang, Bingxuan and others},
  journal={arXiv preprint arXiv:2412.10302},
  year={2024}
}

@misc{llama32,
  author       = {{Meta}},
  title        = {{L}lama 3.2 Model Cards and Prompt Formats},
  howpublished = {\url{https://www.llama.com/docs/model-cards-and-prompt-formats/llama3$\_$2/}},
  year         = {2025},
}

@misc{GLM46v,
      title={GLM-4.5V and GLM-4.1V-Thinking: Towards Versatile Multimodal Reasoning with Scalable Reinforcement Learning}, 
      author={V Team and Wenyi Hong and Wenmeng Yu and Xiaotao Gu and Guo Wang and Guobing Gan and Haomiao Tang and Jiale Cheng and Ji Qi and Junhui Ji and Lihang Pan and Shuaiqi Duan and Weihan Wang and Yan Wang and Yean Cheng and Zehai He and Zhe Su and Zhen Yang and Ziyang Pan and Aohan Zeng and Baoxu Wang and Bin Chen and Boyan Shi and Changyu Pang and Chenhui Zhang and Da Yin and Fan Yang and Guoqing Chen and Jiazheng Xu and Jiale Zhu and Jiali Chen and Jing Chen and Jinhao Chen and Jinghao Lin and Jinjiang Wang and Junjie Chen and Leqi Lei and Letian Gong and Leyi Pan and Mingdao Liu and Mingde Xu and Mingzhi Zhang and Qinkai Zheng and Sheng Yang and Shi Zhong and Shiyu Huang and Shuyuan Zhao and Siyan Xue and Shangqin Tu and Shengbiao Meng and Tianshu Zhang and Tianwei Luo and Tianxiang Hao and Tianyu Tong and Wenkai Li and Wei Jia and Xiao Liu and Xiaohan Zhang and Xin Lyu and Xinyue Fan and Xuancheng Huang and Yanling Wang and Yadong Xue and Yanfeng Wang and Yanzi Wang and Yifan An and Yifan Du and Yiming Shi and Yiheng Huang and Yilin Niu and Yuan Wang and Yuanchang Yue and Yuchen Li and Yutao Zhang and Yuting Wang and Yu Wang and Yuxuan Zhang and Zhao Xue and Zhenyu Hou and Zhengxiao Du and Zihan Wang and Peng Zhang and Debing Liu and Bin Xu and Juanzi Li and Minlie Huang and Yuxiao Dong and Jie Tang},
      year={2025},
      eprint={2507.01006},
      archivePrefix={arXiv},
      primaryClass={cs.CV},
      url={https://arxiv.org/abs/2507.01006}, 
}

@article{wang2025genexam,
  title={GenExam: A Multidisciplinary Text-to-Image Exam},
  author={Wang, Zhaokai and Yin, Penghao and Zhao, Xiangyu and Tian, Changyao and Qiao, Yu and Wang, Wenhai and Dai, Jifeng and Luo, Gen},
  journal={arXiv preprint arXiv:2509.14232},
  year={2025}
}

@inproceedings{li2025chemvlm,
  title={Chemvlm: Exploring the power of multimodal large language models in chemistry area},
  author={Li, Junxian and Zhang, Di and Wang, Xunzhi and Hao, Zeying and Lei, Jingdi and Tan, Qian and Zhou, Cai and Liu, Wei and Yang, Yaotian and Xiong, Xinrui and others},
  booktitle={Proceedings of the AAAI Conference on Artificial Intelligence},
  volume={39},
  pages={415--423},
  year={2025}
}

@article{helber2019eurosat,
  title={Eurosat: A novel dataset and deep learning benchmark for land use and land cover classification},
  author={Helber, Patrick and Bischke, Benjamin and Dengel, Andreas and Borth, Damian},
  journal={IEEE Journal of Selected Topics in Applied Earth Observations and Remote Sensing},
  volume={12},
  number={7},
  pages={2217--2226},
  year={2019},
  publisher={IEEE}
}

@inproceedings{lu2022learn,
    title={Learn to Explain: Multimodal Reasoning via Thought Chains for Science Question Answering},
    author={Lu, Pan and Mishra, Swaroop and Xia, Tony and Qiu, Liang and Chang, Kai-Wei and Zhu, Song-Chun and Tafjord, Oyvind and Clark, Peter and Ashwin Kalyan},
    booktitle={The 36th Conference on Neural Information Processing Systems (NeurIPS)},
    year={2022}
}

@article{fu2025trustgeogen,
  title={Trustgeogen: Scalable and formal-verified data engine for trustworthy multi-modal geometric problem solving},
  author={Fu, Daocheng and Chen, Zijun and Xia, Renqiu and Liu, Qi and Feng, Yuan and Zhou, Hongbin and Zhang, Renrui and Feng, Shiyang and Gao, Peng and Yan, Junchi and others},
  journal={arXiv preprint arXiv:2504.15780},
  year={2025}
}

@inproceedings{huang2025peace,
  title={Peace: Empowering geologic map holistic understanding with mllms},
  author={Huang, Yangyu and Gao, Tianyi and Xu, Haoran and Zhao, Qihao and Song, Yang and Gui, Zhipeng and Lv, Tengchao and Chen, Hao and Cui, Lei and Li, Scarlett and others},
  booktitle={Proceedings of the Computer Vision and Pattern Recognition Conference},
  pages={3899--3908},
  year={2025}
}

@article{wang2025qmirror,
  title={Q-Mirror: Unlocking the Multi-Modal Potential of Scientific Text-Only QA Pairs},
  author={Wang, Junying and Zhang, Zicheng and Shen, Ye and Wu, Yalun and Liang, Yingji and Guo, Yijin and Wen, Farong and Li, Wenzhe and Zhao, Xuezhi and Jia, Qi and others},
  journal={arXiv preprint arXiv:2509.24297},
  year={2025}
}

@misc{xi2025bmmrlargescalebilingualmultimodal,
      title={BMMR: A Large-Scale Bilingual Multimodal Multi-Discipline Reasoning Dataset}, 
      author={Zhiheng Xi and Guanyu Li and Yutao Fan and Honglin Guo and Yufang Liu and Xiaoran Fan and Jiaqi Liu and Jingchao Ding and Wangmeng Zuo and Zhenfei Yin and Lei Bai and Tao Ji and Tao Gui and Qi Zhang and Xuanjing Huang},
      year={2025},
      eprint={2507.03483},
      archivePrefix={arXiv},
      primaryClass={cs.CL},
      url={https://arxiv.org/abs/2507.03483}, 
}

@article{ghadiyaram2015massive,
  title={Massive online crowdsourced study of subjective and objective picture quality},
  author={Ghadiyaram, Deepti and Bovik, Alan C},
  journal={IEEE transactions on image processing},
  volume={25},
  number={1},
  pages={372--387},
  year={2015},
  publisher={IEEE}
}

@article{li2023agiqa,
  title={Agiqa-3k: An open database for ai-generated image quality assessment},
  author={Li, Chunyi and Zhang, Zicheng and Wu, Haoning and Sun, Wei and Min, Xiongkuo and Liu, Xiaohong and Zhai, Guangtao and Lin, Weisi},
  journal={IEEE Transactions on Circuits and Systems for Video Technology},
  volume={34},
  number={8},
  pages={6833--6846},
  year={2023},
  publisher={IEEE}
}

@inproceedings{fang2020perceptual,
  title={Perceptual quality assessment of smartphone photography},
  author={Fang, Yuming and Zhu, Hanwei and Zeng, Yan and Ma, Kede and Wang, Zhou},
  booktitle={Proceedings of the IEEE/CVF conference on computer vision and pattern recognition},
  pages={3677--3686},
  year={2020}
}

@article{zhang2023subjective,
  title={Subjective and Objective Quality Assessment for in-the-Wild Computer Graphics Images},
  author={Zhang, Zicheng and Sun, Wei and Zhou, Yingjie and Jia, Jun and Zhang, Zhichao and Liu, Jing and Min, Xiongkuo and Zhai, Guangtao},
  journal={Transactions on Multimedia Computing, Communications, and Applications (TOMM)},
  year={2023},
  publisher={ACM}
}

\appendix

\section{Data Sources and Statistics}
\label{app:data_sources}

Our image data are sampled from eight open-source datasets, encompassing both interdisciplinary and domain-specific scientific collections. To preserve visual diversity while removing redundancy, we apply perceptual hashing (via HashMAP) to large-scale datasets ($\sim$10K images), filtering out near-duplicates. This yields an initial pool of \textbf{18,702 unique images}.

We adopt a taxonomy inspired by BMMR~\cite{xi2025bmmrlargescalebilingualmultimodal} to categorize images into eight scientific disciplines using MLLMs for coarse-grained classification. To ensure balanced representation across domains during training and evaluation, we uniformly sample up to 2,000 images per category (or all available if fewer). This results in a final working set of \textbf{11,515 images}, which serves as the foundation for both the SIQA-U benchmark and the SIQA-S training dataset. Detailed statistics of data sources are provided in Table~\ref{tab:data_sources}.

\begin{table}[htbp]
\centering
\caption{Summary of data sources in the SIQA benchmark.}
\label{tab:data_sources}
\resizebox{\columnwidth}{!}{%
\begin{tabular}{lccccp{8cm}}
\toprule
Data Source & Whole & Select & Bench & Platform & URL \\
\midrule
BMMR~\cite{xi2025bmmrlargescalebilingualmultimodal}        & 88,991 & 5,000 & 590 & Hugging Face & \url{https://huggingface.co/datasets/guanyu615/BMMR} \\
ChenVLM~\cite{li2025chemvlm}    & 1,689  & 1,689 & 280 & Hugging Face & \url{https://huggingface.co/datasets/Duke-de-Artois/ChemVLM_test_data} \\
EuroSAT~\cite{helber2019eurosat}     & 27,000 & 2,000 & 245 & Github            & \url{https://github.com/phelber/EuroSAT} \\
GenExam~\cite{wang2025genexam}     & 1,000  & 1,000 & 113 & Hugging Face & \url{https://huggingface.co/datasets/OpenGVLab/GenExam} \\
GeoTrust~\cite{fu2025trustgeogen}    & 3,040  & 3,040 & 218 & Hugging Face & \url{https://huggingface.co/datasets/InternScience/GeoTrust} \\
PEACE~\cite{huang2025peace}      & 124    & 124   & 16  & Hugging Face & \url{https://huggingface.co/datasets/microsoft/PEACE} \\
Qmirror~\cite{wang2025qmirror}     & 2,192  & 2,192 & 215 & arXiv        & \url{https://arxiv.org/abs/2509.24297} \\
ScienceQA~\cite{lu2022learn}   & 10,332 & 3,654 & 423 & Hugging Face & \url{https://huggingface.co/datasets/derek-thomas/ScienceQA} \\
\bottomrule
\end{tabular}%
}
\vspace{-1.5em}
\end{table}

\section{The SIQA-U Benchmark: Automated Construction and Validation}
\label{app:siqa_u_construction}

The \textbf{SIQA-U} subset serves as our gold-standard evaluation benchmark. Distinct from traditional datasets, SIQA-U is constructed via a rigorous \textit{automated generation pipeline} grounded in expert-defined dimensions, followed by extensive bias validation to ensure fairness.

\subsection{The SIQA Evaluation Framework}
We propose a four-dimensional framework to holistically evaluate scientific image quality, explicitly accounting for both perceptual interpretability and epistemic integrity:
\begin{description}
    \item[Scientific Validity:] Accuracy of represented facts, principles, and underlying data.
    \item[Scientific Completeness:] Presence of all essential information (labels, scales, context) required for correct interpretation.
    \item[Cognitive Clarity:] Intuitive perceivability and unambiguous interpretability by domain experts.
    \item[Disciplinary Conformity:] Adherence to field-specific conventions regarding visualization style, notation, and color semantics.
\end{description}
To ensure consistent interpretation during automated generation, we instantiated specific prompt templates for each dimension (Figure~\ref{fig:evaluation_dim}).

\begin{figure}[htbp]
\centering
\begin{tcolorbox}[
    width=\linewidth,
    colback=gray!15,
    colframe=gray!60,
    arc=2mm,
    boxrule=0.5pt
]
{\scriptsize 
\begin{verbatim}
Cognitive Clarity = (
    "Scientific images should enhance understanding. "
    "The easier an image is to comprehend, the higher its quality. Consider: "
    "Logical and clear layout improves quality. "
    "Text, labels, and visuals are legible and easy to interpret. "
    "Complex elements are annotated appropriately without excessive clutter.")
Disciplinary Conformity = (
    "Images should align with the conventions of their field."
    "For example: Mathematics uses clean, minimalistic diagrams;"
    "Biology uses colorful, detailed illustrations. Consider: "
    "The visual style and content must match disciplinary norms. "
    "Nomenclature (e.g., chemical names) must follow international standards."
    "No violations of domain-specific rules.")
Scientific Validity = (
    "All content must be factually correct and"
    "non-misleading. Consider: All depicted information must be"
    "accurate and consistent with established knowledge. "
    "No contradictions between visual elements or text. "
    "Data tables, equations, and measurements must be authentic and"
    "not fabricated.")
Scientific Completeness = (
    "Scientific images should present information completely and concisely."
    "Consider: No redundant or repeated information. "
    "Key elements (e.g., axis labels, reactants in chemical reactions)"
    "are not missing. All necessary information is included exactly "
    "once—neither under- nor over-specified.")
\end{verbatim}
}
\end{tcolorbox}
\caption{Prompt for SIQA Dimensions}
\label{fig:evaluation_dim}
\vspace{-1.5em}
\end{figure}

\subsection{Question Typology and Two-Stage Generation Pipeline}
To comprehensively assess MLLM capabilities, we designed three question types targeting complementary aspects of scientific understanding:
\begin{description}
    \item[Yes-or-No:] Binary verification of factual or structural conditions.
    \item[What:] Multiple-choice comprehension of entities, phenomena, or relationships.
    \item[How:] Multiple-choice judgment of visualization appropriateness (quality, clarity, conventions).
\end{description}
Prompt templates for these types are shown in Figure~\ref{fig:question_type}.

\begin{figure}[htbp]
\centering
\begin{tcolorbox}[
    width=\linewidth,
    colback=gray!15,
    colframe=gray!60,
    arc=2mm,
    boxrule=0.5pt
]
{\scriptsize 
\begin{verbatim}
# Question Type Definitions
yes-or-no = (
    "A binary question assessing basic visual understanding, such as the"
    " presence or correctness of fundamental attributes (e.g., labels, layout). Answer"
    "with A.Yes or B.No.")
what = (
    "A multiple-choice question testing comprehension of image content. Include"
    "one correct option and three plausible distractors. Answer using A/B/C/D.")
how = ("A multiple-choice question evaluating perceived image quality "
    "(e.g., clarity, accuracy). Provide four Excellent/Good/Fair/Poor option,"
    " phrased appropriately to the context. Answer using A/B/C/D.")
\end{verbatim}
}
\end{tcolorbox}
\caption{Prompt for Question Types}
\label{fig:question_type}
\vspace{-1.5em}
\end{figure}

\paragraph{Two-Stage Generation Pipeline.}
For each of the 11,515 source images, we employed a two-stage automated pipeline to synthesize high-quality Multiple Choice Questions (MCQs):

\textbf{Stage 1: Dimension-Specific Description Generation.} 
We first prompted a state-of-the-art VLM to generate descriptive content for each of the four quality dimensions. The input included the dimension definition, task instructions, and the image. The model produced dual-level descriptions reflecting both ``high-quality'' and ``low-quality'' instantiations. Prompts for this stage are detailed in Figure~\ref{fig:prompts-description}.

\begin{figure}[htbp]
\centering
\begin{tcolorbox}[
    width=\linewidth,
    colback=gray!15,
    colframe=gray!60,
    arc=2mm,
    boxrule=0.5pt
]
{\scriptsize 
\begin{verbatim}
instruct_prompt_step1 = (
"You are a scientific image quality assessment expert, skilled at"
"analyzing images across specified dimensions."
"You must clearly describe both the strengths and weaknesses of"
"the image, providing specific reasons for each point."
"Your response must strictly follow the required output format."
"Do not add extra paragraphs, titles, or explanations."
)

prompt_step_1 = """Please thoroughly understand the following scientific 
image quality evaluation dimensions and describe the image accordingly:
### Evaluation Dimensions
{evaluate_dim}
### Output Format
"Description of good aspects, reason.
Description of bad aspects, reason.
"""

\end{verbatim}
}
\end{tcolorbox}
\caption{Prompt for Generating Image Descriptions}
\label{fig:prompts-description}
\vspace{-1.5em}
\end{figure}

\textbf{Stage 2: MCQ Synthesis.}
Using the dual-level descriptions, we synthesized MCQ pairs for every combination of dimension and question type. The VLM input comprised the dimension definition, descriptions, question type specification, and the original image. This pipeline produced a total of \textbf{180,834} MCQ pairs. Prompts for synthesis are detailed in Figure~\ref{fig:prompt-question}.

\begin{figure}[htbp]
\centering
\begin{tcolorbox}[
    width=\linewidth,
    colback=gray!15,
    colframe=gray!60,
    arc=2mm,
    boxrule=0.5pt
]

{\scriptsize 
\begin{verbatim}
instruct_prompt_step2 = (
"You are a scientific image assessment question designer, skilled at creating"
"high-quality visual question answering (VQA) questions based on image descriptions"
"and evaluation dimensions. You must generate 1~3 questions that cover diverse aspects"
"of the image. Each question must include options, correct answer (in A/B/C/D format),"
"and explanation. "The output must be valid JSON, strictly conforming to the specified "
"`out_format` structure. Do not add extra fields or formatting."
)
prompt_step_2 = """
    ### Task
    Generate 3–6 high-quality multiple-choice questions based **strictly** on:
  - The provided image description,
  - The given question types,
  - The specified evaluation dimensions,
  - And the underlying content implied by the image itself.
    Each question must assess a distinct aspect of the image content and 
    align with one allowed question type and one evaluation dimension.
    ### Image Description
    {description}
    ### Question Types
    {question_type}
    ### Evaluation Dimensions
    {evaluate_dim}
    ### Output Format
    Output a JSON object with keys "q1", "q2", ..., each containing:
  - "type": a string that exactly matches with the "Question Types" list above.
  - "category": a string that exactly matches with the "Evaluation Dimensions" list above.
  - "question": the question text.
  - "option": a string in the format "A. ... B. ... C. ... D. ..."
  - "answer": the correct choice letter (e.g., "A").
  - "explanation": a brief justification.
    Example structure (do NOT copy the content, only follow the structure):
    {{
        "q1": {{
            "type": "<type from ### Question Type>",
            "category": "<dimension from ### Evaluation Dimensions>",
            "question": "...",
            "option": "A. ... B. ... C. ... D. ...",
            "answer": "A",
            "explanation": "..."
        }},
        ......
    }}
"""
\end{verbatim}
}
\end{tcolorbox}
\caption{Prompt for Generating SIQA-U MCQ Pairs}
\label{fig:prompt-question}
\vspace{-1.5em}
\end{figure}

\subsection{Bias Analysis and Validation}
To verify that our post-hoc rebalancing eliminates self-bias, we conducted a pilot evaluation using four MLLMs, including \textit{doubao-1.5-thinking}, which was utilized during question generation. We sampled 300 items per question type (total 2,100) and evaluated accuracy.

As shown in Table~\ref{tab:bias_analysis}, \textit{doubao-1.5-thinking} performs consistently within the range of independent models (e.g., Yes-or-No: 46.9\% vs. 44.6\%--48.4\%). There is no statistically significant advantage for the generative model, confirming the absence of exploitable self-bias. The consistently lower accuracy on \textit{How} tasks across all models further indicates that difficulty stems from intrinsic cognitive complexity rather than generation artifacts.

\begin{table}[htbp]
\centering
\caption{Accuracy (\%) of four MLLMs on the refined SIQA-U benchmark.}
\label{tab:bias_analysis}
\begin{tabular}{lccc}
\toprule
Model & Yes-or-No & How & What \\
\midrule
qwen-vl-max-no-thinking & 48.4 & 38.2 & 56.4 \\
gpt-4o & 44.6 & 37.0 & 57.4 \\
doubao-1.5-thinking & 46.9 & 39.9 & 56.1 \\
gemmi-thinking & 48.1 & 49.4 & 63.6 \\
\bottomrule
\end{tabular}
\end{table}

\section{The SIQA-S Dataset: Annotation Interface and Quality Control}
\label{app:siqa_s_annotation}

The \textbf{SIQA-S} subset serves as the training dataset for supervised fine-tuning. To ensure high-quality labels, our annotation process relies on two core components: a specialized interface and a strict quality control protocol.

\subsection{The Annotation Interface}
\begin{figure}[htbp]
    \centering
    \includegraphics[width=0.7\linewidth]{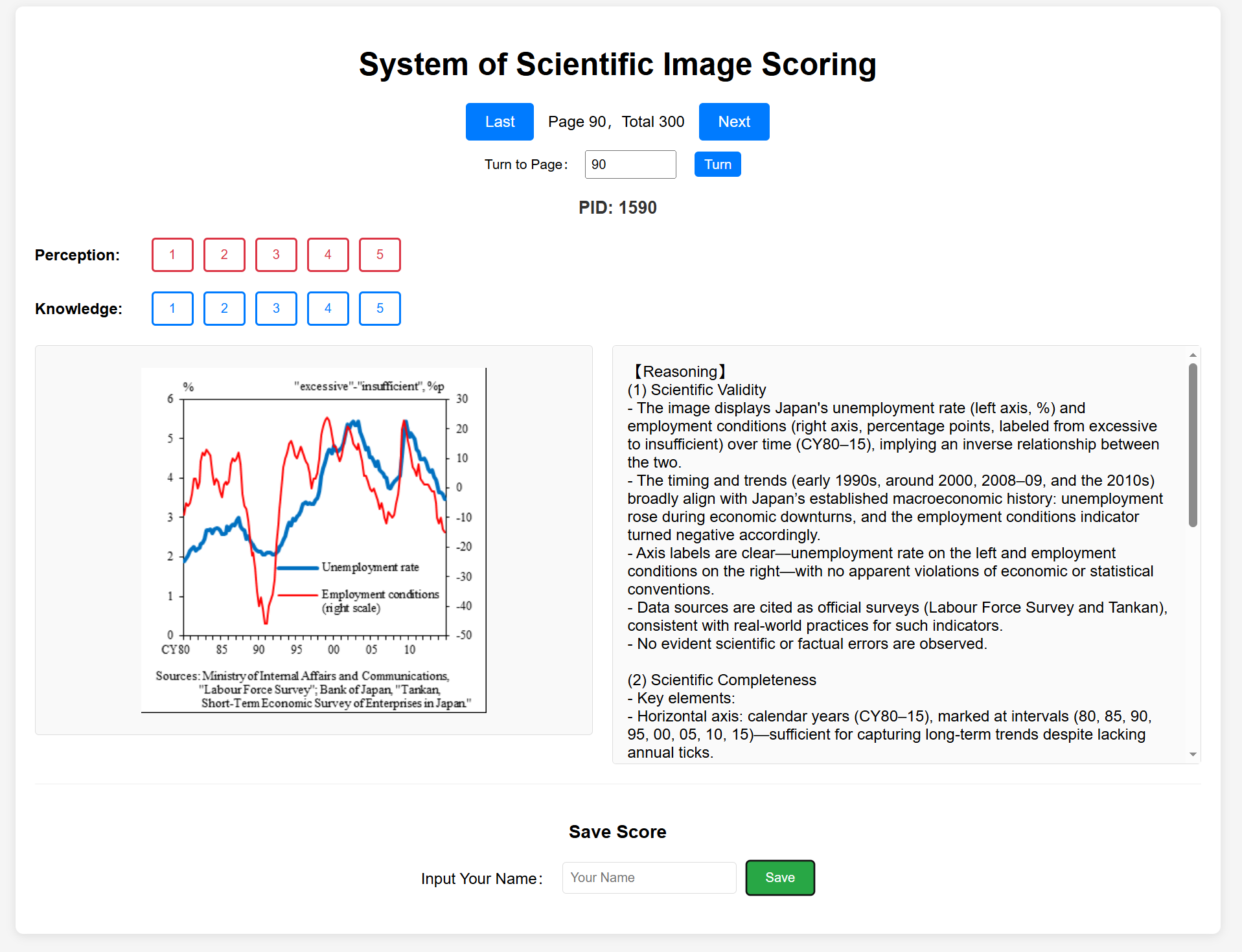}
    \caption{The SIQA-S annotation interface, featuring two-dimensional scoring, a knowledge explanation panel, and efficient navigation controls.}
    \label{fig:annotation interface}
    \vspace{-1.5em}
\end{figure}

As shown in Figure~\ref{fig:annotation interface}, we developed a dedicated interface to facilitate efficient and accurate labeling. The interface features three main components:
\begin{itemize}
    \item \textbf{Two-Dimensional Scoring:} Annotators provide independent ratings for \textit{Perception} (visual quality) and \textit{Knowledge} (scientific correctness) using separate 5-point scales, ensuring distinct evaluation of each aspect.
    \item \textbf{Knowledge Explanation Panel:} A dedicated panel displays detailed textual explanations corresponding to the image content. This provides annotators with necessary reference knowledge to verify scientific facts during the scoring process.
    \item \textbf{Efficient Navigation:} The interface includes convenient page controls (e.g., ``Last'', ``Next'', and direct page input), allowing annotators to quickly browse and label large batches of images without interruption.
\end{itemize}

\subsection{Quality Control Process}
We implemented a multi-stage quality control quality control pipeline covering the pre-, during, and post-annotation stages:

\paragraph{Pre-Annotation: Domain Matching and Testing.}
To minimize noise, we prioritize assigning images to annotators with relevant domain backgrounds (e.g., matching biology images to biologists). Additionally, all annotators must pass a preliminary test annotation task before starting the formal work; only those who meet the quality standards are allowed to proceed, ensuring the rationality of the subsequent labeling.

\paragraph{During Annotation: Reference-Guided Labeling.}
During the labeling process, annotators perform their assessments while referring to the provided knowledge explanations in the interface. This ensures that every score is grounded in verified factual context, thereby guaranteeing the reasonableness of the annotations.

\paragraph{Post-Annotation: Outlier Rejection.}
After data collection, we inspect the annotation results and identify statistical outliers. These inconsistent labels are reviewed and subsequently removed from the dataset to ensure the final training data is clean and reliable.

\section{Experimental Setup and Implementation Details}
\label{app:exp_setup}

To rigorously evaluate diverse architectures against our dual-dimensional framework, we established distinct protocols for data formulation and baseline adaptation. This section first details the conversion of our annotations into Supervised Fine-Tuning (SFT) formats, which serve as the canonical interaction protocol for both training and evaluation. Subsequently, we provide a comprehensive breakdown of experimental configurations for all baselines, explicitly categorizing them into \textit{Zero-Shot} and \textit{Trained} settings based on their architectural constraints and adaptation strategies.

\subsection{SFT Data Conversion Protocols}
To facilitate both the fine-tuning of our proposed model and the prompt-based evaluation of existing MLLMs, we transformed raw SIQA annotations into structured dialogue formats. These conversions define the standard input-output interface for the Understanding (SIQA-U) and Scoring (SIQA-S) tracks.

\subsubsection{SIQA-U Conversion: Constrained MCQ Format}
For the understanding track, we formulate the task as a constrained MCQ problem. As illustrated in Figure~\ref{fig:siqa-u-sft}, the conversion script constructs a dialogue where the system prompt explicitly restricts the output to a single uppercase letter (A, B, C, or D). This standardized format serves two critical purposes:
\begin{enumerate}
    \item \textbf{Fine-Tuning:} It enables MLLMs to learn the task via next-token prediction with strict output constraints.
    \item \textbf{Inference:} It provides a unified prompt template for baseline MLLMs, where responses are parsed to extract the first valid character for accuracy calculation.
\end{enumerate}

\begin{figure}[htbp]
\centering
\begin{tcolorbox}[width=\linewidth, colback=gray!15, colframe=gray!60, arc=2mm, boxrule=0.5pt]
{\scriptsize
\begin{verbatim}
def tran_SIQA_U_OpenAI(data_item, root_path):
    # Validate and extract fields
    image_path = data_item["image_path"].strip()
    question = data_item["question"].strip()
    option = data_item["option"].strip()
    answer = data_item["annotation"].strip().upper()
    
    # Construct constrained prompt
    user_text = (
        "<image>Answer based on the image.\n"
        f"Question: {question}\nChoices: {option}\n"
        "Respond with ONLY one uppercase letter: A, B, C, or D."
    )
    messages = [
        {"role": "system", "content": "You are an expert in scientific image analysis. 
        Output ONLY a single letter."},
        {"role": "user", "content": user_text},
        {"role": "assistant", "content": answer}
    ]
    return {"messages": messages, "images": [os.path.join(root_path, image_path)]}
\end{verbatim}
}
\end{tcolorbox}
\caption{SFT conversion logic for SIQA-U (MCQ Format).}
\label{fig:siqa-u-sft}
\vspace{-1.5em}
\end{figure}

\subsubsection{SIQA-S Conversion: Disentangled Rating Format}
For the scoring track, we decompose the assessment into two orthogonal dimensions: \textit{Perception} and \textit{Knowledge}. As shown in Figure~\ref{fig:siqa-s-sft}, the conversion logic generates two independent dialogue turns for each image, each conditioned on a specialized system prompt (e.g., focusing on ``sharpness'' vs. ``scientific rigor''). This disentanglement is critical for:
\begin{enumerate}
    \item \textbf{Multi-Task Training:} It prevents models from conflating aesthetic appeal with scientific validity during optimization.
    \item \textbf{Dimension-Specific Querying:} It enables precise zero-shot evaluation of baseline MLLMs on each dimension independently.
\end{enumerate}

\begin{figure}[htbp]
\centering
\begin{tcolorbox}[width=\linewidth, colback=gray!15, colframe=gray!60, arc=2mm, boxrule=0.5pt]
{\scriptsize
\begin{verbatim}
def tran_SIQA_S_OpenAI(data_item, root_path):
    terms = ", ".join(RATING_TO_WORD[i] for i in range(1, 6))
    img = os.path.join(root_path, data_item['image_path'].strip())
    
    # Turn 1: Subjective Perception
    perc_prompt = f"Evaluate **Perception Quality** (sharpness, layout). Use: [{terms}]."
    perc_sample = {
        "messages": [
            {"role": "system", "content": perc_prompt + " Respond: Perception: [Word]"},
            {"role": "user", "content": "<image>Rate the subjective quality."},
            {"role": "assistant", "content":
            f"Perception: {float_to_rating_word(data_item['perception_rating'])}"}
        ],
        "images": [img]
    }
    
    # Turn 2: Objective Knowledge
    know_prompt = f"Evaluate **Knowledge Quality** (rigor, completeness). Use: [{terms}]."
    know_sample = {
        "messages": [
            {"role": "system", "content": know_prompt + " Respond: Knowledge: [Word]"},
            {"role": "user", "content": "<image>Rate the objective quality."},
            {"role": "assistant", "content": 
            f"Knowledge: {float_to_rating_word(data_item['knowledge_rating'])}"}
        ],
        "images": [img]
    }
    return perc_sample, know_sample
\end{verbatim}
}
\end{tcolorbox}
\caption{SFT conversion logic for SIQA-S (Disentangled Rating Format).}
\label{fig:siqa-s-sft}
\vspace{-1.5em}
\end{figure}

\subsection{Baseline Experimental Configurations}
We categorized all baselines into \textit{Zero-Shot} and \textit{Trained} settings based on whether they underwent parameter updates using our SIQA-S training set. This distinction ensures a fair comparison between general-purpose models leveraging pre-trained knowledge and those specifically adapted to our scientific domain.

\subsubsection{Group 1: Zero-Shot Evaluation}
Models in this category were evaluated directly using our standardized prompts (Section D.1) without any fine-tuning on the SIQA data. They rely entirely on their pre-existing capabilities or hand-crafted metrics.

\begin{itemize}
    \item \textbf{NIQE:} As a hand-crafted, no-reference metric, NIQE requires no training. It regresses a single continuous score reflecting perceptual naturalness. To align with our framework, we computed correlations between its output and both the \textit{Perception} and \textit{Knowledge} ground truths separately, analyzing its sensitivity to each dimension.
    
    \item \textbf{Q-Align:} Although Q-Align is an MLLM pre-trained on general IQA tasks, it was not fine-tuned on our dataset. We treated it as a zero-shot baseline by feeding it our dimension-specific prompts (Figure~\ref{fig:siqa-s-sft}). For each image, we performed two independent inference passes (one for Perception, one for Knowledge) to obtain dual scores.
    
    \item \textbf{CLIP-IQA:} We extended this method to our setting by constructing two semantically distinct text prototypes: (1) ``A photo with high \textit{perceptual} quality'' and (2) ``A photo with high \textit{scientific knowledge} quality.'' No training was performed; we simply computed the cosine similarity between the image embedding and each text prompt to derive two zero-shot scores.
\end{itemize}

\subsubsection{Group 2: Trained / Fine-Tuned Evaluation}
Models in this category were explicitly trained or fine-tuned on the SIQA-S training set (and SIQA-U for our method) to adapt to our dual-dimensional output space.

\begin{itemize}
    \item \textbf{CLIP-IQA+:} We re-implemented CLIP-IQA+ with a modified prediction head producing a \textbf{2-dimensional vector} $[s_{perc}, s_{know}]$. Following the official training configuration, we fine-tuned both the language projection head and the regression head on our SIQA-S training set using a multi-task loss $\mathcal{L} = \mathcal{L}_{perc} + \mathcal{L}_{know}$. This allows the model to learn domain-specific features for both dimensions.
    
    \item \textbf{HyperIQA:} Being a CNN-based architecture, HyperIQA inherently requires supervised training. We modified its regression head to output \textbf{two scores} and trained the network end-to-end on the joint SIQA-S dataset. The optimization concurrently minimizes errors for both perception and knowledge, yielding two predicted scores during inference.
    
    \item \textbf{SIQA-Judger (Ours):} Our proposed method is a unified MLLM fine-tuned on \textbf{both} SIQA-U (MCQ) and SIQA-S (Rating) tasks using a pure \textbf{language modeling objective}. Distinct from standard IQA methods that rely on regression losses, we utilize a mixed-batch strategy optimizing a unified cross-entropy loss $\mathcal{L}_{total} = \mathcal{L}_{CE}$, where both the correct option letter (SIQA-U) and the textual rating token (SIQA-S) are treated as next-token prediction problems. We deliberately avoid auxiliary regression or ranking losses to ensure simplicity. During inference, the model dynamically switches between generating a discrete class label or a descriptive rating based solely on the system prompt, leveraging shared representations learned from both tasks.
\end{itemize}

\noindent\textbf{Summary of Settings:} Table~\ref{tab:baseline_settings} summarizes the training status and output strategies of all baselines. Note that while Q-Align and CLIP-IQA leverage powerful pre-trained weights, they are evaluated in a strict zero-shot manner relative to our specific scientific domain, whereas CLIP-IQA+, HyperIQA, and SIQA-Judger benefit from explicit adaptation to our data distribution.

\begin{table}[htbp]
\centering
\caption{Summary of baseline experimental settings.}
\label{tab:baseline_settings}
\resizebox{\columnwidth}{!}{%
\begin{tabular}{lccc}
\toprule
Method & Architecture & Training on SIQA & Output Strategy \\
\midrule
NIQE & Hand-crafted & No (Zero-Shot) & 1 Score (Mapped to 2 dims) \\
Other MLLMs & MLLM & No (Zero-Shot) & Dynamic (MCQ or 2 Scores) \\
Q-Align & MLLM & No (Zero-Shot) & 2 Scores (Queried Twice) \\
CLIP-IQA & CLIP-based & No (Zero-Shot) & 2 Scores (Queried Twice) \\
CLIP-IQA+ & CLIP-based & Yes (Fine-tuned) & 2 Scores (Joint Head) \\
HyperIQA & CNN & Yes (Trained) & 2 Scores (Joint Head) \\
SIQA-Judger (Ours) & MLLM & Yes (Multi-task) & Dynamic (MCQ or 2 Scores) \\
\bottomrule
\end{tabular}%
}
\vspace{-1.5em}
\end{table}

\end{document}